\documentclass{amsart}

\usepackage[utf8]{inputenc}
\usepackage[a4paper, margin=2cm]{geometry}
\usepackage{amssymb,latexsym,enumitem,bbm,xcolor}
\usepackage[noadjust]{cite}
\usepackage{setspace}
\usepackage{color}
\usepackage{hyperref}
\usepackage{tikz}
\usepackage{amsmath} 
\usepackage{amsthm}
\usepackage{amsfonts}
\usepackage{pgfplots}
\pgfplotsset{compat=1.18}
\usetikzlibrary{arrows}
\usepackage{graphicx}
\usepackage{subcaption}
\usepackage{float} 

\numberwithin{equation}{section}
\theoremstyle{plain} 
\newtheorem{theorem}{Theorem}[section]
\newtheorem{corollary}[theorem]{Corollary}
\newtheorem{lemma}[theorem]{Lemma}
\newtheorem{proposition}[theorem]{Proposition}

\theoremstyle{definition} 
\newtheorem{definition}[theorem]{Definition}

\newtheorem{example}[theorem]{Example}

\theoremstyle{remark} 
\newtheorem{remark}[theorem]{Remark}

\allowdisplaybreaks
\title{\textbf{Deep convolutional neural networks based on the special affine Fourier transform}}

\date{} 
	\author{Radha Ramakrishnan}
	\address{Department of Mathematics, Indian Institute of Technology Madras, 600036. India}
	\email{radharam@iitm.ac.in}
	\author{Amritha Varrier}
	\address{Department of Mathematics, Indian Institute of Technology Madras, 600036. India}
	\email{ma24d032@smail.iitm.ac.in}
	
\begin{document}


\begin{abstract}
   The aim of this paper is to investigate the deep convolutional neural networks (DCNN) associated with the special affine Fourier transform (SAFT). We show that the corresponding network can be
made asymptotically $\mathcal{A}$-translation invariant by an appropriate choice of pooling
factors. We further prove that the feature extractor is deformation-insensitive for $\mathcal{A}$-bandlimited functions and cartoon functions. We also study the propagation of
energy through the layers. Finally, we define the discrete SAFT and apply it for the digit classification on the MNIST dataset. In particular, we evaluate the accuracy, training behavior and sensitivity of the model to the parameters of
the affine transform against conventional methods. 
\end{abstract}

\maketitle

\section{\textbf{Introduction}}

The special affine Fourier transform (SAFT) was originally introduced 
by  Abe and Sheridan \cite{article1} in the context of analyzing certain 
operations on optical wave functions. Formally, the SAFT is defined by 
the integral operator
\begin{equation}
F_A f(\omega) = \frac{1}{\sqrt{|b|}} 
\int_{\mathbb{R}} f(t)\, e^{\frac{\pi i}{b}\left(at^2 + 2pt - 2\omega t 
+ d\omega^2 + 2(bq - dp)\omega\right)}\, dt, \quad \omega \in \mathbb{R},
\tag{1.1}
\end{equation}
where \(A = \{a, b, c, d, p, q\}\) denotes a set of real parameters 
subject to the constraints \(ad - bc = 1\) and \(b \neq 0\). The 
transform (1.1) is intrinsically linked to the special affine linear 
transformation of phase space,
\begin{equation}
\begin{pmatrix} t' \\ \omega' \end{pmatrix} 
= \begin{pmatrix} a & b \\ c & d \end{pmatrix} 
\begin{pmatrix} t \\ \omega \end{pmatrix} 
+ \begin{pmatrix} p \\ q \end{pmatrix}.
\tag{1.2}
\end{equation}
Owing to the conditions imposed on the parameters \(a, b, c, d\), the 
matrix appearing in (1.2) is an element of the special linear group 
\(SL(2, \mathbb{R})\); consequently, the affine transformation (1.2) 
is realized by elements of the inhomogeneous linear group
\[
ISL(2, \mathbb{R}) = \left\{ 
\begin{pmatrix} M & v \\ 0 & 1 \end{pmatrix} 
: M \in SL(2, \mathbb{R}),\; v \in \mathbb{R}^2 \right\}.
\]
This group-theoretic structure provides the justification for the 
terminology ``special affine Fourier transform'' assigned to (1.1). 
The action of \(SL(2, \mathbb{R})\) on the time--frequency plane, 
together with its relation to quadratic Fourier transforms, has been 
thoroughly investigated in the literature; for a comprehensive 
treatment, the reader is referred to \cite{MR2827662}.\\

A number of classical integral transforms arise as particular cases 
of the SAFT. Specifically, the parameter set 
\(A = \{0, 1, -1, 0, 0, 0\}\) recovers the ordinary Fourier transform, 
whereas \(A = \{0, -1, 1, 0, 0, 0\}\) yields its inverse. The choice 
\(A = \{\cos\theta, \sin\theta, -\sin\theta, \cos\theta, 0, 0\}\) 
corresponds to the fractional Fourier transform, and 
\(A = \{1, \lambda, 0, 1, 0, 0\}\) produces the Fresnel transform.\\

The SAFT has been studied extensively in the context of harmonic analysis and its applications. In \cite{MR3958693}, Bhandari and Zayed studied shift invariant spaces associated with the SAFT using chirp modulation and generalised Shannon sampling theorem in the context of SAFT domain.~Biswas et al. in \cite{MR4983614}, introduced the translation operator \(T^{A}\) associated with the SAFT and
established analogues of Wendel's theorem and Wiener's theorem in the
SAFT setting. Further, they obtained characterisation results for the system of $A$-translates $\{T_k \phi:k \in \mathbb{Z} \}$ with the generator $\phi \in L^2(\mathbb{R})$ to form a frame sequence and a Riesz sequence.  Subsequently, ~Biswas et al. in \cite{MR4458766}, studied several fundamental properties of
the SAFT in connection with Fourier analysis and time-frquency analysis and introduced the modulation space \(M_{A}^{r,s}\) in
connection with it. They also established Hörmander multiplier theorem and Littlewood--Paley theorem
associated with the SAFT. These investigations laid the groundwork for extending translation-based
signal analysis beyond the classical Fourier setting. This
motivated us for the development of deep convolutional architectures built upon
such generalized transforms. \\

Mallat~\cite{MR2957703} introduced a deep convolutional neural network (DCNN) based on the wavelet transform, built from cascaded convolutions and the modulus nonlinearity, and without pooling; the resulting architecture is translation invariant. This construction was subsequently generalized by Wiatowski and B\"olcskei~\cite{MR3766318} by replacing wavelets with semi-discrete frames and allowing general choices of nonlinearities and pooling operators across layers.  In~\cite{MR3958697}, Czaja and Li developed feature extractors that integrate Mallat's scattering framework with time--frequency (Gabor) representations by introducing \,uniform covering frames\,, a class that contains many semi-discrete Gabor systems. In~\cite{MR4666815}, Nicola et al. studied how the scattering architecture interacts with the deformation regularity measured on the H\"older scale, identifying stability thresholds: stability may persist for certain deformations, while instability can arise at lower regularity. Later, Chua et al.~\cite{MR4643737} generalized finite-depth wavelet scattering by expressing it in terms of $L^q(\mathbb{R}^n)$ ($1 \leq q \leq 2$) norms of cascades of (continuous or dyadic) wavelet transforms composed with contractive nonlinearities. In ~\cite{MR4702505}, Chew et al. proposed a unified framework for geometric scattering on measure spaces, encompassing earlier settings (e.g., compact Riemannian manifolds without boundary and undirected graphs) and extending to directed and signed graphs as well as manifolds with boundary. Biswas et al. \cite{MR4983277} studied the DCNN using the fractional Fourier transform and studied the $\Theta$-translation invariance property in this setting.\\

Beyond invariance, two additional properties are central to the analysis and validation of feature extractors: stability with respect to small input deformations and the layer-wise behavior of energy. Wiatowski and B\"olcskei~\cite{MR3766318} showed how structural features of the input signal are reflected in the resulting feature vectors, and proved deformation stability for band-limited inputs. However, many practically relevant signals exhibit sharp or curved discontinuities and are not band-limited. Motivated by this, Grohs et al.~\cite{GrohsWB16} established deformation stability of the feature extractor for cartoon functions (introduced by Donoho in~\cite{article2}), which explicitly model these structural characteristics. Later, Wiatowski et al. \cite{MR3819343} established conditions for energy conservation for a broad class of DCNN-based feature extractors and characterized corresponding feature-map energy decay rates. In \cite{MR4924777}, F\"uhr and Getter  investigated energy decay in CNNs and established that, for wavelet and wavelet-like scattering transforms acting on generic square-integrable signals, energy decay can be arbitrarily slow. Moreover, they demonstrated that this slow-decay behavior occurs on dense subsets of $L^2(\mathbb{R})$, indicating that rapid energy decay is generally an unstable signal property. They complemented these findings with sufficient conditions that yield fast (up to exponential) energy decay for generalized Sobolev spaces tailored with the frequency localization of the underlying filter bank. In \cite{MR4135431}, Zou and Lerman proved that energy is conserved at every layer of the CNN in the graph setting. Wiatowski et al.~\cite{pmlr-v48-wiatowski16} analyzed DCNNs based on the discrete Fourier transform and derived both local and global sensitivity bounds with respect to deformations and translations.\\

\subsection{Our contributions}
 One of the questions mentioned in the future research directions of \cite{MR4983277} is to investigate networks in connection with general transforms such as the linear canonical transform and the Fresnel transform which are particular cases of the more general SAFT and also to investigate the performance, advantages, and disadvantages of these networks
compared to other available networks. In this paper, we wish to study DCNN in connection with the more general special affine Fourier transform (SAFT). Towards this end, we first generalize the SAFT to higher dimensions,
denoting it by \(\mathcal{F}_{\mathcal{A}}\), which is constructed as the
tensor product of one-dimensional SAFTs. Subsequently, we extend the
\(\mathcal{A}\)-translation operator \(T^{\mathcal{A}}\), the
\(\mathcal{A}\)-modulation operator \(M^{\mathcal{A}}\), and the chirp
modulation operator \(C_{\frac{a}{b}}\) to higher dimensions corresponding to
\(\mathcal{F}_{\mathcal{A}}\). Then, we construct a semi-discrete system associated with the SAFT via the operator \(T^{\mathcal{A}}\). Let \(\Lambda\) be a countable index set and $\{g_{\lambda}:\lambda\in\Lambda\}\subset L^1(\mathbb{R}^n)\cap L^2(\mathbb{R}^n)$. We derive characterisation results that specify when the semi-discrete system \(\{T_s^{\mathcal{A}}\check{g}_{\lambda}: s\in\mathbb{R}^n,\,\lambda\in\Lambda\}\) forms a semi-discrete frame for \(L^2(\mathbb{R}^n)\), where \(\check f(x)=f(-x)\). We formulate the DCNN in the SAFT framework by making use of the above semi-discrete frame, $\mathcal{A}$-convolution, non-linearities and pooling factors. The DCNN associated with the SAFT need not, in general, be translation invariant. Nevertheless, it can be rendered asymptotically invariant under the \(\mathcal{A}\)-translations through an appropriate choice of pooling factors. Moreover, we provide Python implementations for functions on \(\mathbb{R}\) and \(\mathbb{R}^2\), using semi-discrete frames generated by $C_{\frac{a}{b}}D_s\psi$ for various choices of the parameters $a$, $b$, and $s$, where $\psi$ denotes the Haar wavelet. All these results are discussed in Section 3 of this paper.\\

In Section~4, we introduce the class of \(\mathcal{A}\)-band-limited functions and show that, for this class, the feature extractor becomes asymptotically insensitive to deformations. We also investigate deformation insensitivity for cartoon functions. In Section~5, we derive several results on how energy is conserved in each layer and how it propagates across layers. We introduce the operator $D_{X,p}^\mathcal{A}$ associated with an invertible operator $X$ and examine its interaction with the scattering propagator. In particular, we prove that the energy remainder is invariant under changes of dilation, provided the semi-discrete frame is dilated inversely.\\

In the concluding section of this paper, we introduce the discrete special affine Fourier transform (DSAFT) along with a
numerical implementation on the MNIST dataset. This discretisation makes the
SAFT amenable to numerical computation, and we subsequently use it as a
representation mechanism within a neural architecture trained on MNIST
handwritten digits. Empirically, the proposed discrete SAFT-based model attains higher classification accuracy than a conventional fully connected network (FCN), suggesting that the discrete SAFT provides a more effective
representation than standard dense layers.
\vspace{0.5cm}
  
The following definition is used in the sequel of the paper. Let $\mathcal{H}$ denote a separable Hilbert space.  
\begin{definition}
A sequence $\{f_k:k\in\mathbb{N}\}$ in $\mathcal{H}$ is called a \emph{frame} for $\mathcal{H}$ if there exist constants
\[
0<C_1\le C_2<\infty
\]
such that
\begin{equation}\label{eq:frame}
C_1\|f\|^2
\le
\sum_{k=1}^{\infty}\left|\langle f,f_k\rangle\right|^2
\le
C_2\|f\|^2,
\qquad \forall\, f\in\mathcal{H}.
\end{equation}

If only the right hand side inequality in \eqref{eq:frame} holds, then
 $\{f_k:k\in\mathbb{N}\}$ is called a \emph{Bessel sequence}. If
$C_1=C_2$, then  $\{f_k:k\in\mathbb{N}\}$ is called a
\emph{tight frame}, in particular, if $C_1=C_2=1$, it is called a \emph{Parseval frame}. The constants $C_1$ and $C_2$ are referred to as the
\emph{lower} and the \emph{upper frame bounds}, respectively.
\end{definition}

For a study of frames, we refer to \cite{christensen2016introduction} and \cite{heil2011basis}.
\vspace{0.5cm}
\section{\textbf{Special affine Fourier transform in higher dimensions}}
\vspace{0.5cm}
The special affine Fourier transform of an $f \in L^1(\mathbb{R})$ is defined as
\begin{equation}
\mathcal{F}_A(f)(\omega)
=
\frac{1}{\sqrt{| b|}}
\int_{\mathbb{R}}
f(t)
\exp\left\{
\frac{\pi i}{b}
\left(
at^2
-2t(\omega-p)
-2\omega(dp-bq)
+d\omega^2
\right)
\right\}\,dt,
\qquad \omega\in\mathbb{R}.
\end{equation}
Now, we shall define the SAFT in higher dimensions.

\begin{definition}
    Let $a,b,c,d,p,q\in \mathbb{R}^n$ and $f \in L^1(\mathbb{R}^n)$. Then the \emph{special affine Fourier transform (SAFT)} of $f$, denoted by $\mathcal{F_A}(f)$, is defined as \\
    \begin{equation}
    \mathcal{F_A}(f)
:=   \mathcal{F}_{A_1}^{t_1} \cdot \mathcal{F}_{A_2}^{t_2}\cdots \mathcal{F}_{A_n}^{t_n} ,
    \end{equation}
where $ \mathcal{F}_{A_j}^{t_j}$ is the one-dimensional special affine Fourier transform of $f$ in the $t_j$ variable with parameters $A_j=[a_j,b_j,c_j,d_j,p_j,q_j],$ where $1 \leq j \leq n$. Moreover, we have for $\omega\in\mathbb{R}^n$,
\begin{equation}
 \mathcal{F_A}(f)(\omega)
= \frac{1}{\sqrt{|b_1 \cdot b_2 \cdots b_n|}} \int_{\mathbb{R}^n}f(t)
\exp\bigg( \displaystyle \sum_{j=1}^n
\frac{\pi i}{b_j}
\left(
a_jt_j^2
-2t_j(\omega_j-p_j)
-2\omega_j(d_jp_j-b_jq_j)
+d_j\omega_j^2
\right)
\bigg)\,dt. \quad
\end{equation}
\end{definition}
\begin{definition}
    For $a,b \in \mathbb{R}^n$ and $f: \mathbb{R}^n \rightarrow \mathbb{C}$, a measurable function, the  \emph{chirp modulation operator}, denoted by $C_{\frac{a}{b}}$, is defined as\\
    \begin{equation}
        C_{\frac{a}{b}}f(t):=\exp\bigg(\pi i\displaystyle \sum_{j=1}^n \frac{ a_j t_j^2}{b_j} \bigg)f(t), \quad t \in \mathbb{R}^n.
    \end{equation}
\end{definition}
Using this operator, we can write the SAFT of $f$ in terms of the classical Fourier transform of $f$, namely $\hat{f}$: \\
\begin{equation}
 \mathcal{F_A}(f)(\omega)=\frac{\eta_\mathcal{A}(\omega)}{\sqrt{|b_1 \cdot b_2 \cdots b_n|}}(C_{\frac{a}{b}}f) \widehat(\frac{\omega-p}{b}), \quad \omega \in \mathbb{R}^n, 
\end{equation} where $\eta_\mathcal{A}(\omega)=\exp\bigg( \displaystyle \sum_{j=1}^n \frac{\pi i}{b_j}
\left( -2\omega_j(d_jp_j-b_jq_j)
+d_j\omega_j^2
\right)
\bigg).$

\begin{example}
As an illustrative example, Fig.~1 displays the one-dimensional special affine Fourier transforms of the box function \(\chi_{[1,\,2]}\) (top) and the Gaussian \(e^{-\frac{x^2}{2}}\) (bottom), evaluated for different values of parameters. Because this transform yields complex-valued outputs, we plot both the magnitude and the real/imaginary components.
\end{example}

\begin{example}
In Fig.~2 we show, from top to bottom, the special affine Fourier transform (SAFT) for the 2 dimensional function $\chi_{[1,2]} \otimes \chi_{[1,2]}$ (top) and the Gaussian \(e^{-\frac{x_1^2+x_2^2}{2}}\) (bottom) in \(\mathbb{R}^2\) for $A_j=(1, 1, 0, 1, 3, 0)$ for $j=1,2$. 

\end{example}

\begin{example}
    Let $f(x,y)=\chi_{[-1/2,1/2]}(x)\chi_{[-1/2,1/2]}(y).$ Then $C_{-\frac{a}{b}}f(t_1,t_2)= \exp\bigg(- \pi i \big(\frac{a_1t_1^2}{b_1}+\frac{a_2t_2^2}{b_2}\big)\bigg)f(t_1,t_2).$ Thus, $\mathcal{F_A}(C_{-\frac{a}{b}}f)$ is given by\\
\[
    \mathcal{F_A}(C_{-\frac{a}{b}}f(\omega_1,\omega_2))=\frac{1}{\sqrt{|b_1 \cdot b_2|}}\eta_\mathcal{A}(\omega) \operatorname{sinc}(\frac{\omega_1-p_1}{b_1})\operatorname{sinc}(\frac{\omega_2-p_2}{b_2}),
\]
    
    where 
    $\operatorname{sinc}(x) =
\begin{cases}
\dfrac{\sin(\pi x)}{\pi x}, & x \neq 0\\[6pt]
1 \qquad \quad, & x = 0
\end{cases}.$

If $A_j=A=[a,b,c,d,p,q]$ for $j=1,2,$ then 
 $C_{-\frac{a}{b}}f(t_1,t_2)= \exp\big(- \pi i \frac{a}{b}||t||^2 \big)f(t_1,t_2)$ and 
 $\mathcal{F_A}(C_{-\frac{a}{b}}f(\omega_1,\omega_2))=\frac{1}{|b|}\eta_\mathcal{A}(\omega) \operatorname{sinc}(\frac{\omega_1-p}{b})\operatorname{sinc}(\frac{\omega_2-p}{b})$.
\end{example}

The translation and modulation operators associated with the one-dimensional special aﬃne Fourier transform have already been introduced in \cite{MR4458766} and  \cite{MR4983614}. Now, we define the translation, modulation and convolution operators associated with SAFT for higher dimensions.
\begin{definition}
    Let $s \in \mathbb{R}^n$ and $f: \mathbb{R}^n \rightarrow \mathbb{C}$ be a measurable function. Then the \emph{$\mathcal{A}$-translation} of $f$ by $s$, denoted by $T_s^\mathcal{A}(f)$, is defined as \\
    \begin{equation}
        T_s^\mathcal{A}(f)(t):=\exp\bigg(-2 \pi i \displaystyle \sum_{j=1}^n \frac{s_j(t_j-s_j) a_j}{b_j}\bigg) f(t-s), \quad t\in \mathbb{R}^n.
    \end{equation}
\end{definition}

\begin{figure}[!t]
    \centering
    \begin{subfigure}{0.3\textwidth}
        \centering
        \includegraphics[width=0.8\linewidth]{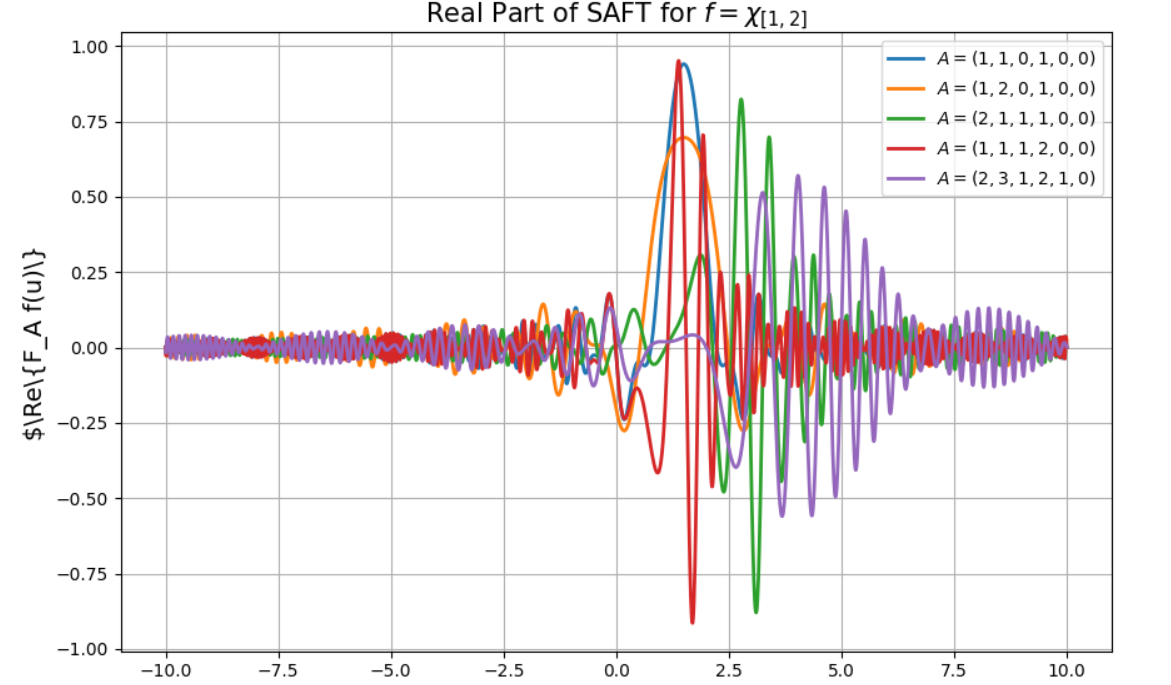}
        \caption{Real part}
        \label{fig:example-1-row1a}
    \end{subfigure}
    \begin{subfigure}{0.3\textwidth}
        \centering
        \includegraphics[width=0.8\linewidth]{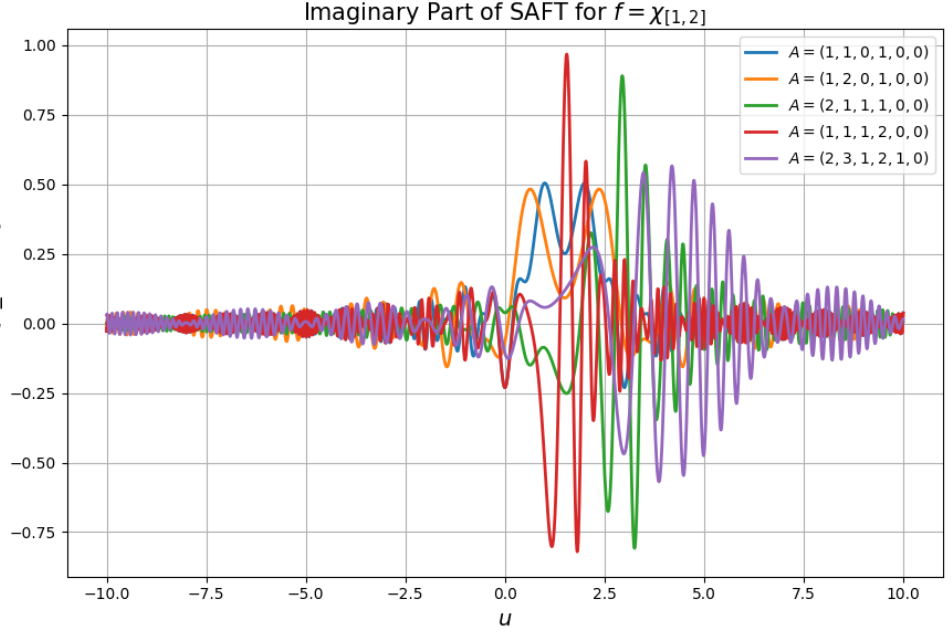}
        \caption{Imaginary part}
        \label{fig:example-1-row1b}
    \end{subfigure}
    \begin{subfigure}{0.3\textwidth}
        \centering
        \includegraphics[width=0.8\linewidth]{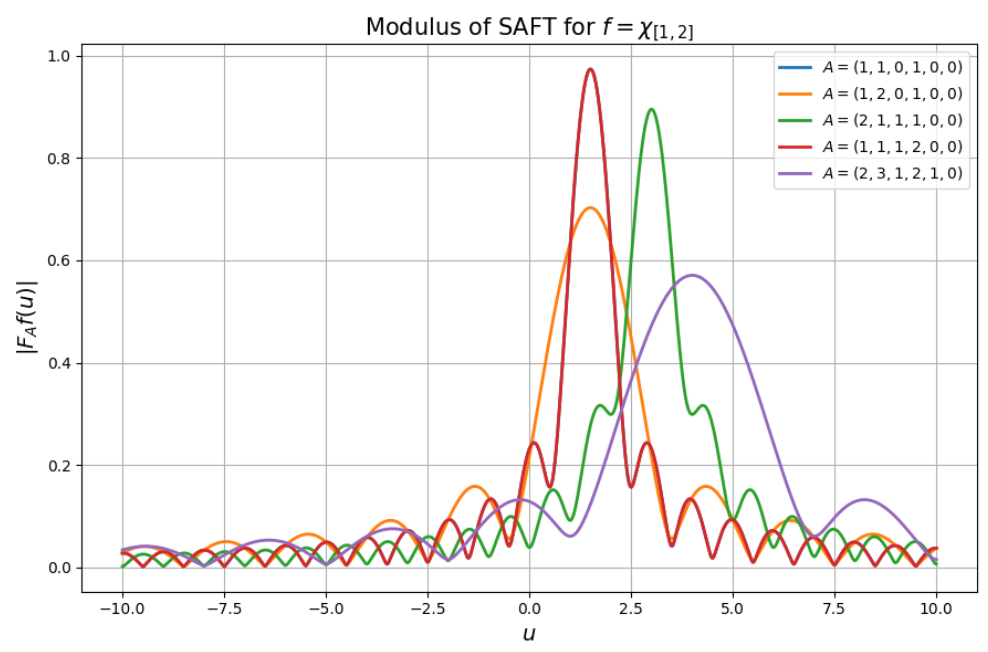}
        \caption{Modulus}
        \label{fig:example-1-row1c}
    \end{subfigure}
    
    \vspace{1em} 
    
    \begin{subfigure}{0.3\textwidth}
        \centering
        \includegraphics[width=0.8\linewidth]{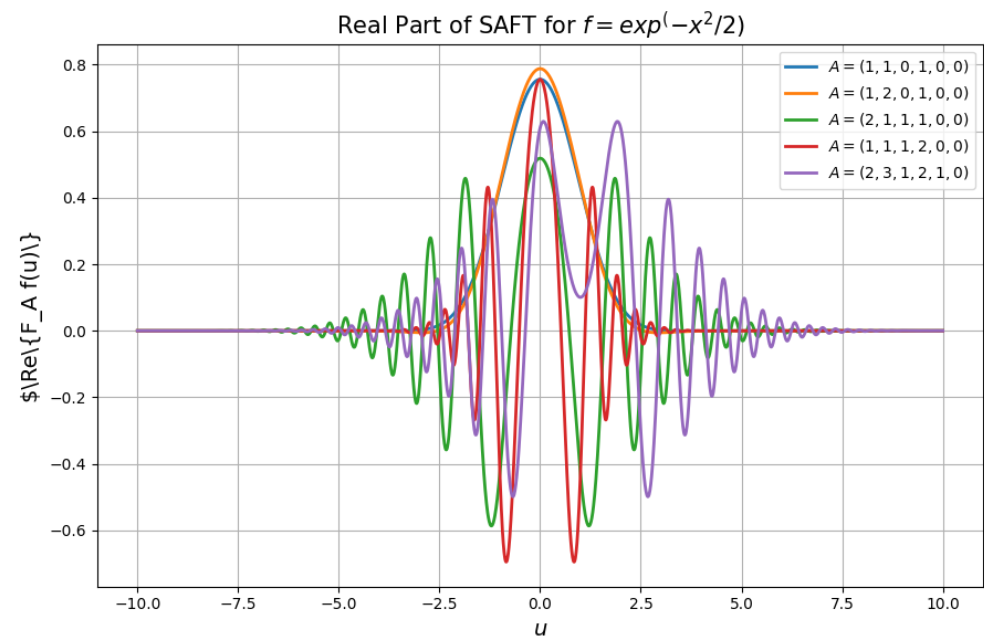}
        \caption{Real part}
        \label{fig:example-1-row2a}
    \end{subfigure}
    \begin{subfigure}{0.3\textwidth}
        \centering
        \includegraphics[width=0.8\linewidth]{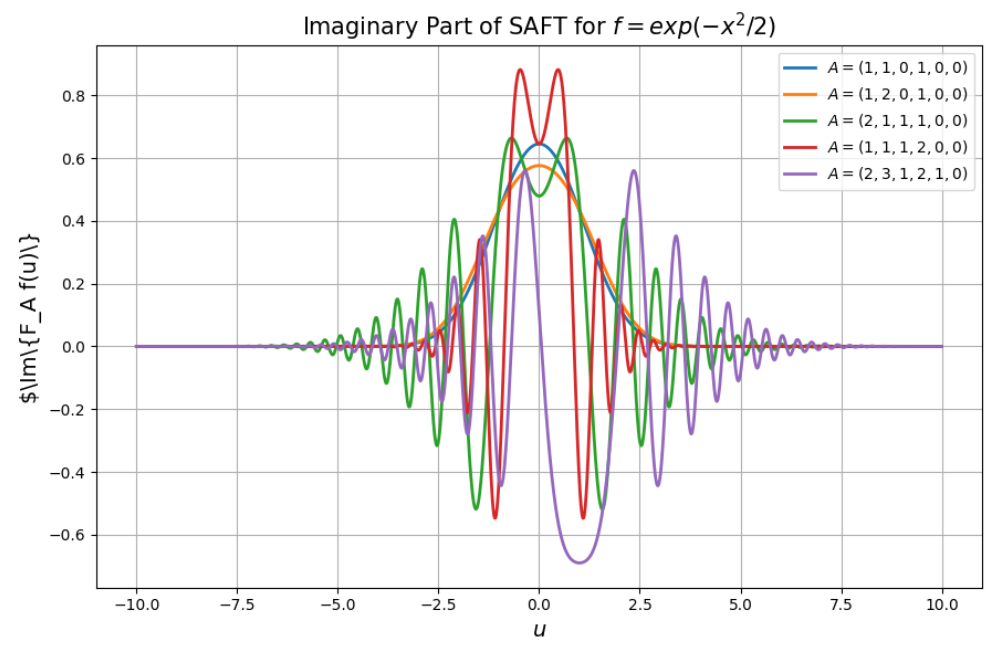}
        \caption{Imaginary part}
        \label{fig:example-1-row2b}
    \end{subfigure}
    \begin{subfigure}{0.3\textwidth}
        \centering
        \includegraphics[width=0.8\linewidth]{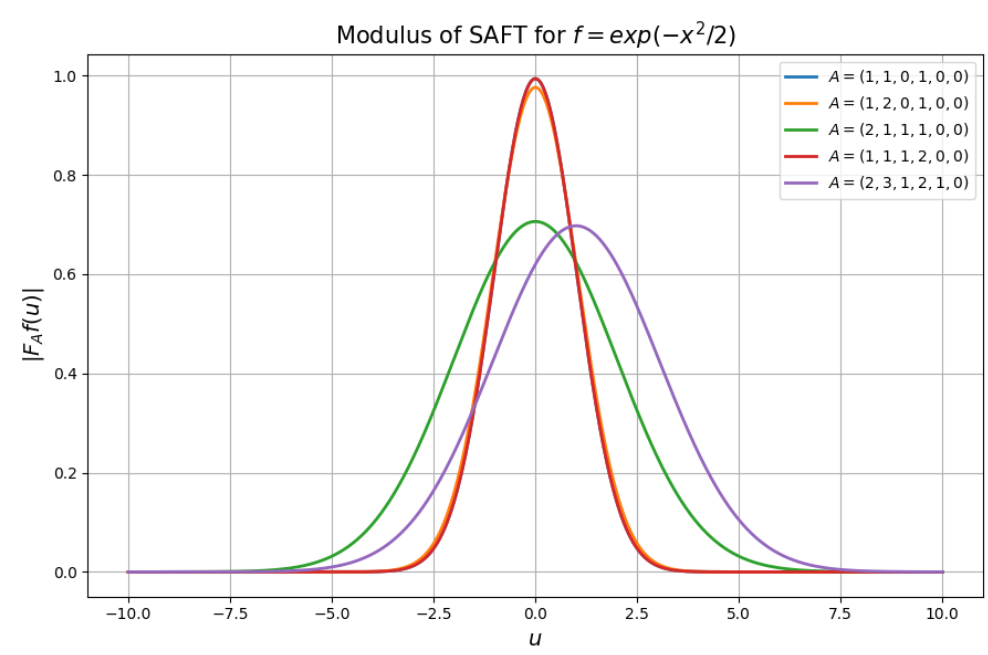}
        \caption{Modulus}
        \label{fig:example-1-row2c}
    \end{subfigure}
    
    \caption{Special affine Fourier transforms of the characteristic function $\chi_{[1,2]}$ and the Gaussian function $e^{-\frac{x^2}{2}}$ in $\mathbb{R}$}
    \label{fig:example-1}
\end{figure}
\begin{figure}[!t]
    \centering
    \begin{subfigure}{0.3\textwidth}
        \centering
        \includegraphics[width=0.8\linewidth]{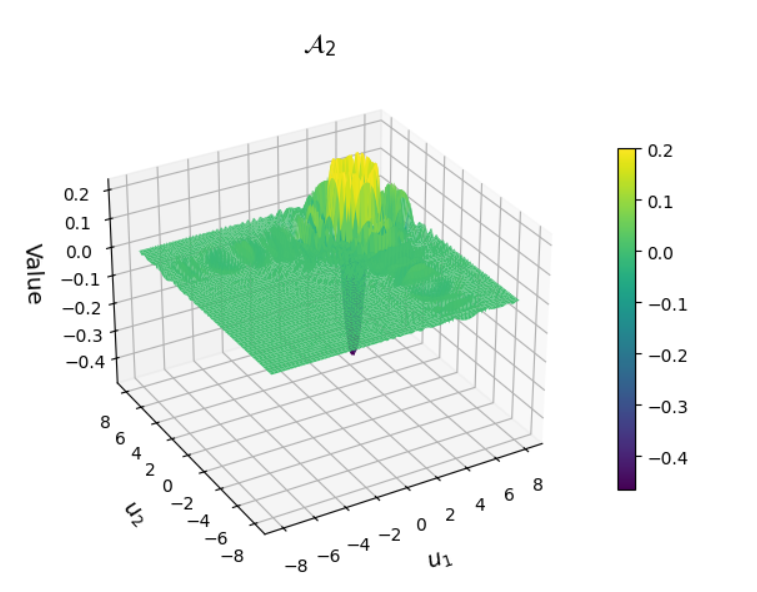}
        \caption{Real part}
        \label{fig:example-2-row1a}
    \end{subfigure}
    \begin{subfigure}{0.3\textwidth}
        \centering
        \includegraphics[width=0.8\linewidth]{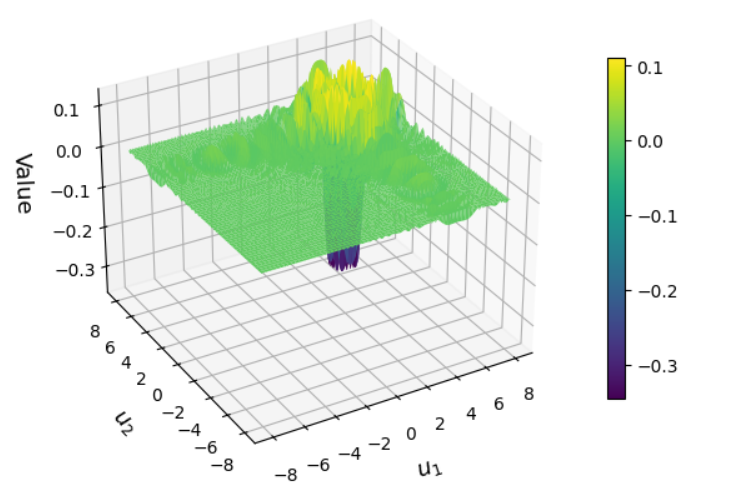}
        \caption{Imaginary part}
        \label{fig:example-2-row1b}
    \end{subfigure}
    \begin{subfigure}{0.3\textwidth}
        \centering
        \includegraphics[width=0.8\linewidth]{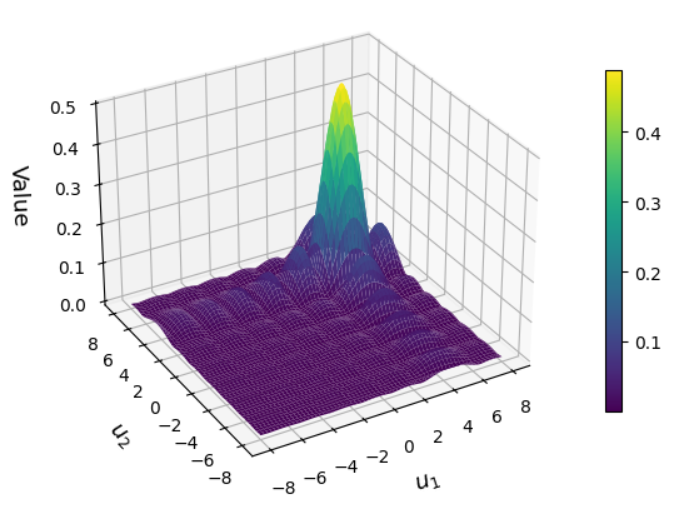}
        \caption{Modulus}
        \label{fig:example-2-row1c}
    \end{subfigure}
    
    \vspace{1em} 
    
    \begin{subfigure}{0.3\textwidth}
        \centering
        \includegraphics[width=0.8\linewidth]{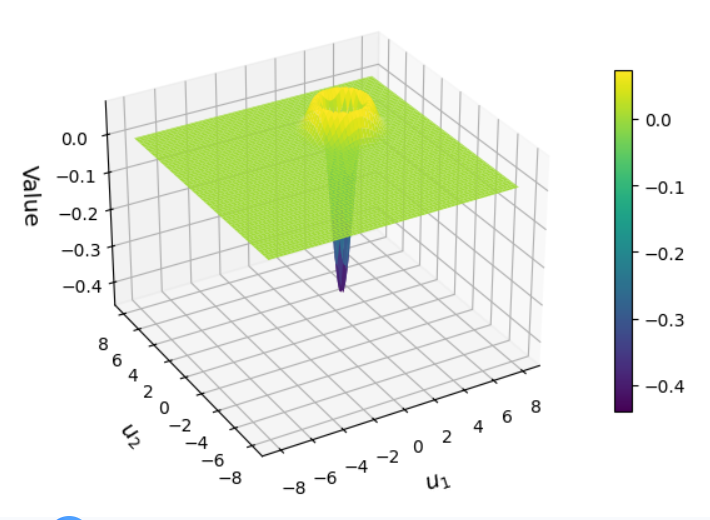}
        \caption{Real part}
        \label{fig:example-2-row2a}
    \end{subfigure}
    \begin{subfigure}{0.3\textwidth}
        \centering
        \includegraphics[width=0.8\linewidth]{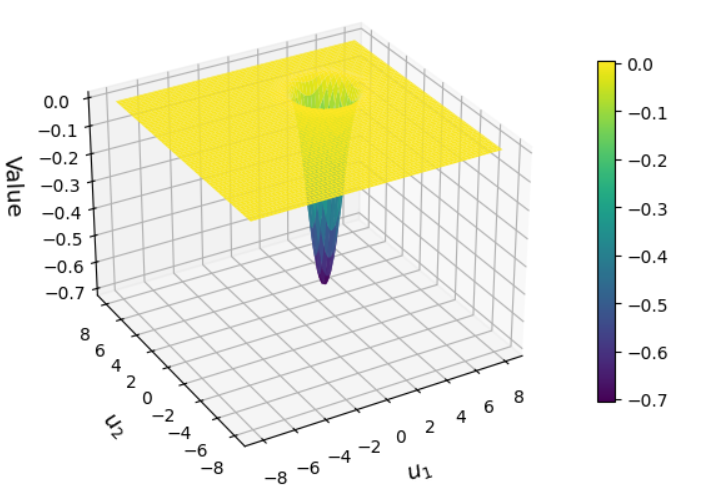}
        \caption{Imaginary part}
        \label{fig:example-2-row2b}
    \end{subfigure}
    \begin{subfigure}{0.3\textwidth}
        \centering
        \includegraphics[width=0.8\linewidth]{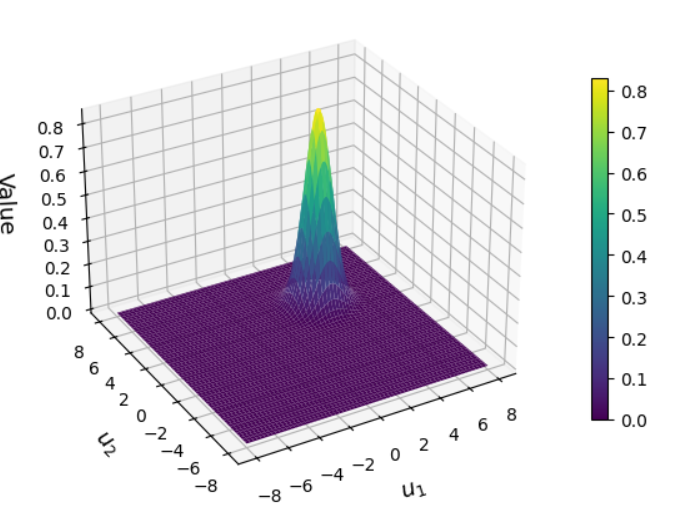}
        \caption{Modulus}
        \label{fig:example-2-row2c}
    \end{subfigure}
    
    \caption{Special affine Fourier transform of the characteristic function $\chi_{[1,2]} \otimes \chi_{[1,2]}$ and the Gaussian $e^{-\frac{x_1^2+x_2^2}{2}}$ in $\mathbb{R}^2$ }
    \label{fig:example-2}
\end{figure}

\begin{remark}
The relationship between the $\mathcal{A}$-translation operator and the classical translation operator using the chirp modulation operator is given below
 \\
\begin{equation}\label{eq:chirp trans}
    C_{\frac{a}{b}}(T_s^\mathcal{A}f)=\exp\bigg( \pi i \displaystyle \sum_{j=1}^n \frac{s_j^2 a_j}{b_j}\bigg) T_s(C_{\frac{a}{b}}f).
    \end{equation}
\end{remark}

\begin{definition}
    Let $s \in \mathbb{R}^n$ and $f: \mathbb{R}^n \rightarrow \mathbb{C}$ be a measurable function. Then, the \emph{$\mathcal{A}$-modulation} of $f$, denoted by  $M_s^\mathcal{A}(f)$, is defined as\\
    \begin{equation}
      M_s^\mathcal{A}(f)(t):=\exp \bigg(\pi i \displaystyle \sum_{j=1}^n \frac{a_j s_j^2-2p_js_j+2s_jt_j}{b_j} \bigg)f(t), \quad t \in \mathbb{R}^n.
    \end{equation}
\end{definition}
\begin{remark}
 Like in the classical case, the SAFT of translation operator is the modulation with opposite sign of the SAFT.
    \begin{equation} \label{eq:mod trans}
        \mathcal{F_A}(T_s^\mathcal{A}f)=M_{-s}^\mathcal{A}(\mathcal{F_A}f).
    \end{equation}
\end{remark}

\begin{example}
    In Figure 4., we display the real and imaginary part of $\mathcal{A}$-modulation of  $f(t_1,t_2)=\frac{\sin t_1}{t_1}\frac{\sin t_2}{t_2}$ in $\mathbb{R}^2.$
\end{example}

\begin{definition}
   For $f,g \in \mathbb{R}^n$, the \emph{$\mathcal{A}$-convolution of $f$ and $g$}, denoted by $f \ast_{\mathcal{A}}g$, is defined as follows.
   \begin{equation}
       (f \ast_{\mathcal{A}}g)(x):= \frac{1}{\sqrt{|b_1 \cdot b_2 \cdots b_n|}} \int_{\mathbb{R}^n}f(s)T_s^\mathcal{A}g(x) ds, \quad  x\in \mathbb{R}^n.
   \end{equation}
   Then, we have
   \begin{equation} \label{eq:chirp conv}
       C_{\frac{a}{b}}(f \ast_{\mathcal{A}}g)=\frac{1}{\sqrt{|b_1 \cdot b_2 \cdots b_n|}}( C_{\frac{a}{b}}f \ast  C_{\frac{a}{b}}g), 
   \end{equation}
   where $f \ast g$ is the classical convolution on $\mathbb{R}^n$.
\end{definition}

\vspace{0.5cm}
\begin{remark}
    For $f,g \in \mathbb{R}^n$, we have
    \begin{equation}
        \mathcal{F_A}(f \ast_{\mathcal{A}}g)(\omega)=\overline{\eta_\mathcal{A}(\omega)}\mathcal{F_A}(f)(\omega)\mathcal{F_A}(g)(\omega), \quad \omega \in \mathbb{R}^n.
    \end{equation}
\end{remark}

\begin{figure}[!t]
    \centering
    \begin{subfigure}{0.3\textwidth}
        \centering
        \includegraphics[width=0.8\linewidth]{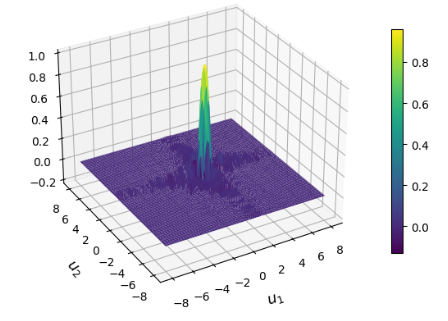}
        \caption{Real part}
        \label{fig:example-3-row1a}
    \end{subfigure}
    \begin{subfigure}{0.3\textwidth}
        \centering
        \includegraphics[width=0.8\linewidth]{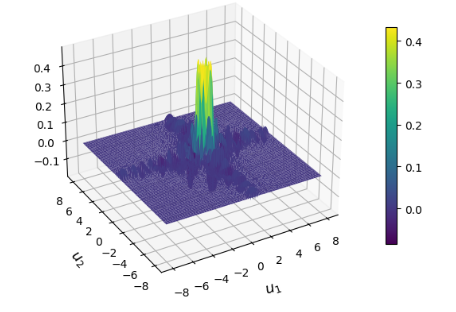}
        \caption{Imaginary part}
        \label{fig:example-3-row1b}
    \end{subfigure}
    \begin{subfigure}{0.3\textwidth}
        \centering
        \includegraphics[width=0.8\linewidth]{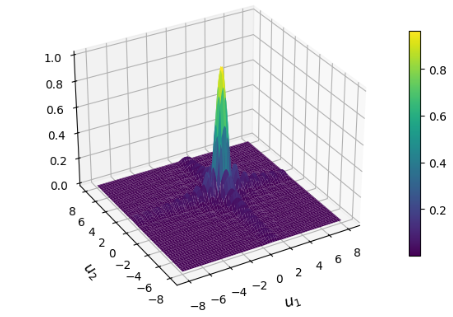}
        \caption{Modulus}
        \label{fig:example-3-row1c}
    \end{subfigure}
    
    \vspace{1em} 
    
    \begin{subfigure}{0.3\textwidth}
        \centering
        \includegraphics[width=0.8\linewidth]{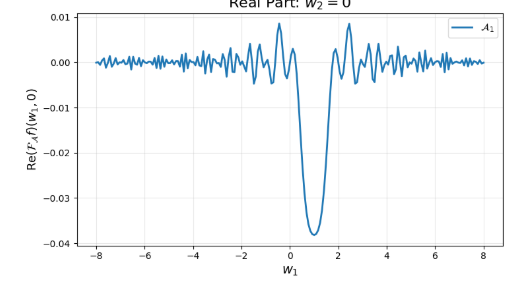}
        \caption{Real part}
        \label{fig:example-3-row2a}
    \end{subfigure}
    \begin{subfigure}{0.3\textwidth}
        \centering
        \includegraphics[width=0.8\linewidth]{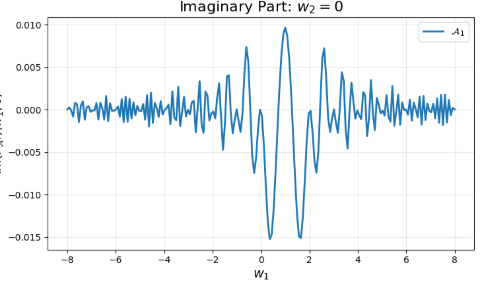}
        \caption{Imaginary part}
        \label{fig:example-3-row2b}
    \end{subfigure}

    \caption{Real part, imaginary part and modulus of the special affine Fourier transform of $C_{-\frac{a}{b}}f$ with $A_j=[2,1,1,1,1,0]$ for $j=1,2$(top) and  the cross section of the real and imaginary part
of the special affine Fourier transform of $C_{-\frac{a}{b}}f$ with $\omega_2=0$}
    \label{fig:example-3}
\end{figure}

\vspace{0.75cm}
\section{\textbf{Semi-discrete SAFT systems and deep CNNs}}
\vspace{0.5cm}
In this section, we introduce semi-discrete systems that are intrinsically linked to the special affine Fourier transform $\mathcal{F}_\mathcal{A}$ through the $\mathcal{A}$-translation operator. We then use these systems to construct a deep convolutional neural network that operates in the SAFT domain. Finally, we investigate the invariance of this network under $\mathcal{A}$-translations.

For clarity, we adopt the following notation for use in the proposition that follows. The idempotent operator on \(L^p(\mathbb{R}^n)\), $1 \leq p \leq \infty$ is defined as
\[
 \check{f}(x) := f(-x).
\]

\begin{definition}
Let $\{g_\lambda:\lambda\in\Lambda\}\subseteq
L^1(\mathbb{R}^n)\cap L^2(\mathbb{R}^n)$, where $\Lambda$ is a countable index set. Define
\[
\Psi_\Lambda^\mathcal{A}
=
\{T_s^\mathcal{A}\check{g_\lambda}: s\in\mathbb{R}^n,\ \lambda\in\Lambda\}.
\]

The collection $\Psi_\Lambda^\mathcal{A}$ is called a
\emph{semi-discrete frame} for $L^2(\mathbb{R}^n)$ if there exist constants
$0<C_1\le C_2<\infty$ such that
\[
C_1\|f\|^2
\le
\sum_{\lambda\in\Lambda}
\int_{\mathbb{R}^n}
\left|
\left\langle
f,T_s^\mathcal{A}\check{g_\lambda}
\right\rangle
\right|^2
\,ds
\le
C_2\|f\|^2,
\qquad
\forall\,f\in L^2(\mathbb{R}^n).
\]
 The functions
$g_\lambda, \lambda\in\Lambda$ are called the
\emph{atoms} (or generators) of the semi-discrete frame
$\Psi_\Lambda$.
\end{definition}

Now, we have a characterization of a system of semi-discrete frames with the help of SAFT. 
\begin{proposition}\label{prop:equivalence}
The semi-discrete system consisting of the \(\mathcal{A}\)-translates of the reflected functions, namely $\quad$
\{$T_s^\mathcal{A}\check{g_\lambda}:  s \in \mathbb{R}^n, \lambda \in \Lambda \}$
constitutes a semi-discrete frame for \(L^2(\mathbb{R}^n)\) with frame bounds \(C_1\) and \(C_2\) if and only if the following two-sided energy inequality holds in the  special affine Fourier domain:
\[
\frac{C_1}{ |b_1 \cdot b_2 \cdots b_n|} \leq \sum_{\lambda \in \Lambda} |\mathcal{F}_\mathcal{A}(g_\lambda)(\omega)|^2 \leq \frac{C_2}{|b_1 \cdot b_2 \cdots b_n|} ,
\]
for almost every \(\omega \in \mathbb{R}^n\).
\end{proposition}

\begin{proof}
    \allowdisplaybreaks

    Consider
    \begin{align*}
        \langle T_s^\mathcal{A}\check{g_\lambda}, f \rangle
        &= \int_{\mathbb{R}^n}T_s^\mathcal{A}\check{g_\lambda}(t) \overline{f(t)} \,dt \\
        &= \int_{\mathbb{R}^n} \exp\bigg(-2 \pi i \sum_{j=1}^n \frac{s_j(t_j-s_j) a_j}{b_j}\bigg) g_\lambda(s-t) \overline{f(t)} \,dt \\
        &= \exp \bigg(\pi i \sum_{j=1}^n \frac{a_j s_j^2}{b_j} \bigg) \int_{\mathbb{R}^n} \exp \bigg(\pi i\sum_{j=1}^n \frac{ a_j (s_j-t_j)^2}{b_j} \bigg) g_\lambda(s-t) \\
        &\qquad\qquad\qquad\qquad\qquad\qquad\times \exp \bigg(-\pi i\sum_{j=1}^n \frac{ a_j t_j^2}{b_j} \bigg)\overline{f(t)}\,dt \\
        &= \exp\bigg(\pi i \sum_{j=1}^n \frac{a_j s_j^2}{b_j} \bigg) \int_{\mathbb{R}^n} C_{\frac{a}{b}} g_\lambda(s-t) \, C_{-\frac{a}{b}}\overline{f(t)}\,dt.
    \end{align*}

    But we have $C_{\frac{a}{b}}^2=C_{\frac{2a}{b}}$ and $\overline{C_{-\frac{a}{b}}}=C_{\frac{a}{b}}$. Hence
    \begin{align*}
        \langle T_s^\mathcal{A}\check{g_\lambda}, C_{-\frac{a}{b}}^2f \rangle
        &= \exp \bigg(\pi i \sum_{j=1}^n \frac{a_j s_j^2}{b_j} \bigg) \int_{\mathbb{R}^n} C_{\frac{a}{b}} g_\lambda(s-t) \, C_{-\frac{a}{b}}\overline{C_{-\frac{a}{b}}^2f(t)}\,dt\\
        &= \exp \bigg(\pi i \sum_{j=1}^n \frac{a_j s_j^2}{b_j} \bigg) \int_{\mathbb{R}^n} C_{\frac{a}{b}} g_\lambda(s-t)\, C_{\frac{a}{b}}\overline{f(t)}\,dt\\
        &= \exp \bigg(\pi i \sum_{j=1}^n \frac{a_j s_j^2}{b_j} \bigg) \big(C_{\frac{a}{b}} g_\lambda \ast C_{\frac{a}{b}} \overline{f}\big)(s)\\
        &= \exp \bigg(\pi i \sum_{j=1}^n \frac{a_j s_j^2}{b_j} \bigg) \sqrt{|b_1 \cdot b_2 \cdots b_n|} \, C_{\frac{a}{b}} \big(g_\lambda \ast_\mathcal{A}\overline{f}\big)(s).
    \end{align*}

    Suppose $\{T_s^\mathcal{A}\check{g_\lambda}, s \in \mathbb{R}^n, \lambda \in \Lambda $\} is a semi-discrete frame. Then, there exist $C_1, C_2>0$ such that\\
    \begin{equation}
    C_1\|f\|^2
\le
\sum_{\lambda\in\Lambda}
\int_{\mathbb{R}^n}
\left|
\left\langle
T_s^\mathcal{A}\check{g_\lambda},f
\right\rangle
\right|^2
\,ds
\le
C_2\|f\|^2,
\qquad
\forall\,f\in L^2(\mathbb{R}^n).
\end{equation}\\

Hence\\
\begin{equation} \label{eq:chirp}
\begin{aligned}
C_1\|C_{-\frac{a}{b}}^2f\|^2
&\le
\sum_{\lambda\in\Lambda}
\int_{\mathbb{R}^n}
\left|
\left\langle
T_s^\mathcal{A}\check{g_\lambda},
C_{-\frac{a}{b}}^2f
\right\rangle
\right|^2
\,ds \\
&\le
C_2\|C_{-\frac{a}{b}}^2f\|^2 .
\end{aligned}
\end{equation}
    But $||f||=||C_{\frac{a}{b}}f||=||C_{-\frac{a}{b}}f||=|| \overline{f}||$. Hence, it follows from \eqref{eq:chirp} that \\
 
   \begin{equation}
       C_1\|f\|^2
\le
\sum_{\lambda\in\Lambda}
\int_{\mathbb{R}^n}
|b_1 \cdot b_2 \cdots b_n| | (g_\lambda \ast_\mathcal{A}f)(s)|^2 ds \le
C_2\|f\|^2.
  \end{equation}
  Since SAFT is a unitary operator and    $|\mathcal{F_A}(f \ast_{\mathcal{A}}g)(\omega)|=|\mathcal{F_A}(f)(\omega)\mathcal{F_A}(g)(\omega)|$, we have  

   \begin{equation}
       C_1\|\mathcal{F_A}(f)\|^2
\le
\sum_{\lambda\in\Lambda}
\int_{\mathbb{R}^n}
|b_1 \cdot b_2 \cdots b_n||\mathcal{F_A}(f)(\omega)|^2|\mathcal{F_A}(g_\lambda)(\omega)|^2 d\omega \le
C_2\|\mathcal{F_A}(f)\|^2
  \end{equation}.\\
 This is true for all $f \in L^2(\mathbb{R}^n)$. Thus, we have
  \begin{equation}
    \frac{C_1}{ |b_1 \cdot b_2 \cdots b_n|} \leq \sum_{\lambda \in \Lambda} |\mathcal{F}_\mathcal{A}(g_\lambda)(\omega)|^2 \leq \frac{C_2}{|b_1 \cdot b_2 \cdots b_n|} ,\quad a.e \quad\omega \in \mathbb{R}^n
  \end{equation}
  We get the converse by retracing the steps back.

\end{proof}

\begin{definition}
Let $\{\Lambda_k\}_{k\in\mathbb{N}}$ be a sequence of countable index sets. For each $k\in\mathbb{N}$, let
\[
\Psi_k^\mathcal{A}=
\left\{
T_s^\mathcal{A} \check{g}_{\lambda_k}
:\;
s\in\mathbb{R}^n,\;
\lambda_k\in\Lambda_k
\right\}
\]
be a semi-discrete frame for $L^2(\mathbb{R}^n)$. Further, let
\[
\mathcal{M}_k^\mathcal{A}:L^2(\mathbb{R}^n)\to L^2(\mathbb{R}^n),\qquad
P_k^\mathcal{A}:L^2(\mathbb{R}^n)\to L^2(\mathbb{R}^n)
\]
be Lipschitz continuous operators satisfying
\[
\mathcal{M}_k^\mathcal{A} f=P_k^\mathcal{A} f=0,
\qquad
\text{whenever } f=0,\; k\in\mathbb{N}.
\]
Then, the sequence of triples
\[
\Omega^\mathcal{A}:=
\left\{
\left(
\Psi_k^\mathcal{A},
\mathcal{M}_k^\mathcal{A},
P_k^\mathcal{A}
\right)
:k\in\mathbb{N} \right\}
\]
is called an \emph{$\mathcal{A}$-module sequence}.
\end{definition}

The classical dilation operator is not compatible with the special affine Fourier transform (SAFT). Consequently, we introduce a modified dilation operator that is consistent with the SAFT framework. This operator will be used throughout the subsequent developments of the network architecture.

    \begin{figure}[!t]
    \centering
    \begin{subfigure}{0.3\textwidth}
        \centering
        \includegraphics[width=0.8\linewidth]{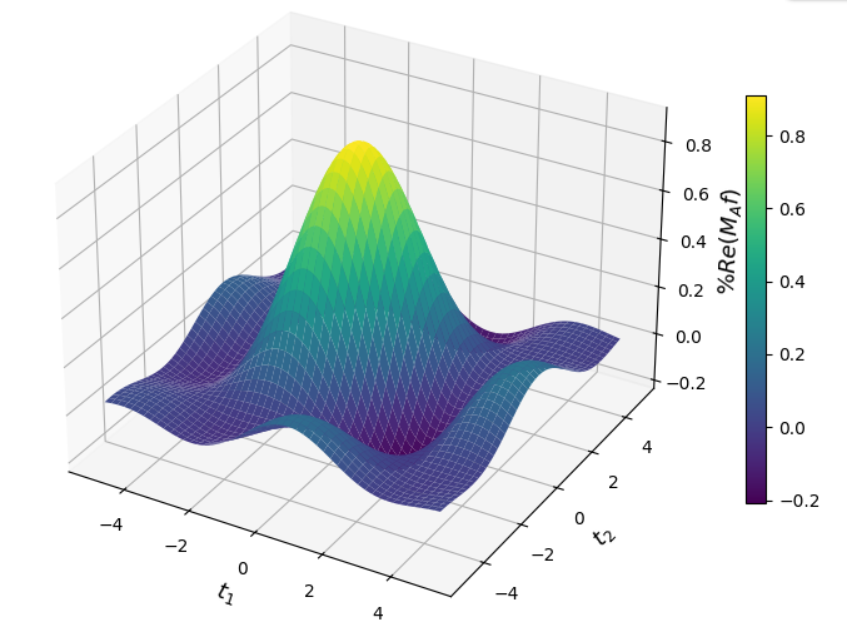}
        \caption{Real part}
        \label{fig:example-4-row1a}
    \end{subfigure}
    \begin{subfigure}{0.3\textwidth}
        \centering
        \includegraphics[width=0.8\linewidth]{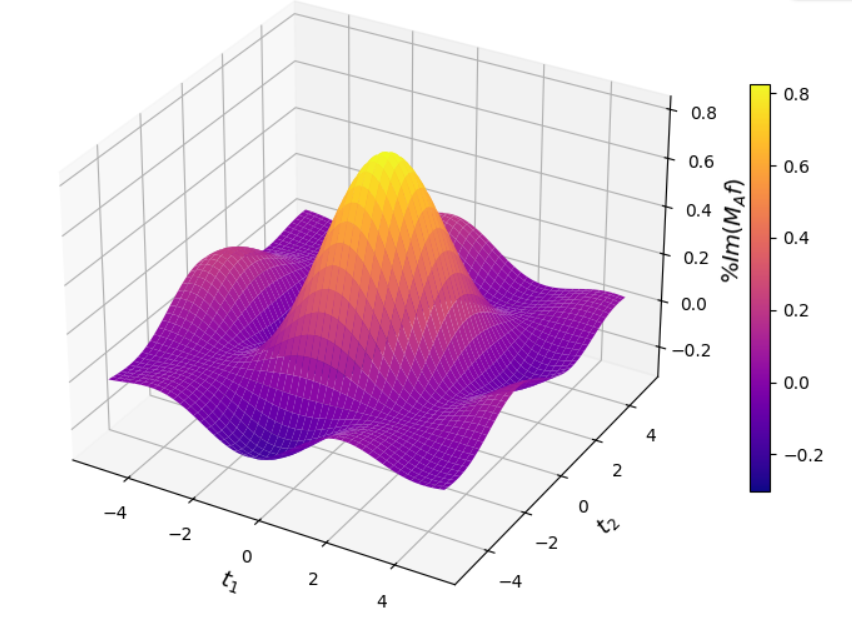}
        \caption{Imaginary part}
        \label{fig:example-4-row1b}
    \end{subfigure}

    \vspace{1em} 

    \caption{The real part and  imaginary part of the $\mathcal{A}$-modulation of $f(t_1,t_2)=\frac{\sin t_1}{t_1}\frac{\sin t_2}{t_2}$ with $a_j=2,b_j=17$ for $j=1,2.$}
    \label{fig:example-4}
\end{figure}

\begin{definition}
    For $s>0$, the \emph{$\mathcal{A}$-dilation operator}, $D_s^\mathcal{A}$ is defined as follows.
    \begin{equation}
        D_s^\mathcal{A}=C_{-\frac{a}{b}} D_s C_{\frac{a}{b}}.
    \end{equation}
\end{definition}
\begin{example}
   In Figure 5., we display the real and imaginary part of $\mathcal{A}$-dilation of  $f(t_1,t_2)=\frac{\sin t_1}{t_1}\frac{\sin t_2}{t_2}$ in $\mathbb{R}^2.$ 
\end{example}
 Let $s_k:s_k>1$ be a sequence of positive real numbers that represent the pooling factors. In order to simplify the notation in the subsequent discussion, the dilation operator $D_{s_k}^{\mathcal{A}}$ will be simply denoted by $D_k^{\mathcal{A}}$.

We next introduce the operator associated with each layer of the network, which serves as the fundamental component of the feature extraction process.

\begin{definition}
Let $\Omega^\mathcal{A}=
\left\{
\left(
\Psi_k^\mathcal{A},
\mathcal{M}_k^\mathcal{A},
P_k^\mathcal{A}
\right)
:k\in\mathbb{N} \right\}$ be an $\mathcal{A}$-module sequence. Suppose that
$
\{g_{\lambda_k}:\lambda_k\in\Lambda_k\}$
is the collection of atoms generating the semi-discrete frame $\Psi_k^\mathcal{A}$ and  $s_k>1$ is the pooling factor corresponding to the $k$-th network layer. Then, the operator associated with the $k$-th layer
\[
U_k^\mathcal{A}:\Lambda_k\times L^2(\mathbb{R}^n)\rightarrow L^2(\mathbb{R}^n),
\]
is defined by
\[
U_k^\mathcal{A}(\lambda_k,f)
=
U_k^\mathcal{A}[\lambda_k]f
:=
D_k^\mathcal{A}
P_k^\mathcal{A}
\mathcal{M}_k^\mathcal{A}
\left(f\star_\mathcal{A} g_{\lambda_k}\right).
\]
\end{definition}
Let $B_k^\mathcal{A}$ be an upper frame bound for the semi-discrete frame $\left\{
T_s^\mathcal{A}\check{g}_{\lambda_k}
:\,
s\in\mathbb{R}^n,\,
\lambda_k\in\Lambda_k
\right\},$ $L_k^\mathcal{A}$ and $R_k^\mathcal{A}$, the Lipschitz constants corresponding to the operators $\mathcal{M}_k^\mathcal{A}$ and $P_k^\mathcal{A}$, respectively. Then,

\begin{equation}
\begin{aligned}
\|U_k^\mathcal{A}[\lambda_k]f\|_2
&\le
\left\|
P_k^\mathcal{A}
\mathcal{M}_k^\mathcal{A}
\left(f\star_\mathcal{A} g_{\lambda_k}^{\mathcal{A}}\right)
\right\|_2 \le
B_k^\mathcal{A}
(R_k^\mathcal{A})^2
(L_k^\mathcal{A})^2
\|f\|_2.
\end{aligned}
\label{eq:Uk-bound}
\end{equation}
Hence, $U_k^\mathcal{A}$ is a well defined operator. 

For every integer $k\geq1$, define
\[
\Lambda^k=\Lambda_1\times\Lambda_2\times\cdots\times\Lambda_k.
\]
An element
\[
q=(\lambda_1,\lambda_2,\ldots,\lambda_k)\in\Lambda^k
\]
is called a \emph{path} of length $k$. The length of an empty path $q=\phi$ is zero.

The operator introduced above can now be extended from a single network layer to an arbitrary path.
\begin{definition}
Let
$\mathcal{Q}
=
\displaystyle \bigcup_{k=0}^{\infty}\Lambda^k.$
Define the operator
$
U^\mathcal{A}:\mathcal{Q}\times L^2(\mathbb{R}^n)
\longrightarrow
L^2(\mathbb{R}^n)$
by
\[
U^\mathcal{A}[q]f
=
U^\mathcal{A}[\lambda_1,\ldots,\lambda_k]f
:=
U_k^\mathcal{A}[\lambda_k]
\cdots
U_2^\mathcal{A}[\lambda_2]
U_1^\mathcal{A}[\lambda_1]f,
\]
where
$
q=(\lambda_1,\ldots,\lambda_k)\in\Lambda^k
$ and $f\in L^2(\mathbb{R}^n).$\\
We also define $U^\mathcal{A}[\phi]f=f.$
\end{definition}
Using the estimate obtained in \eqref{eq:Uk-bound}, it follows that the operator $U^\mathcal{A}$ is well defined. In fact, for any path
$q=(\lambda_1,\ldots,\lambda_k)\in\Lambda^k$ and $f,g \in L^2(\mathbb{R}^n)$, we have
\begin{equation}
\|U^\mathcal{A}[q]f - U^\mathcal{A}[q]g \|_2
\le
\prod_{j=1}^{k}
\left(
B_j^\mathcal{A}
(L_j^\mathcal{A})^2
(R_j^\mathcal{A})^2
\right)
\|f-g\|_2. \label{eq:lipschitz}
\end{equation}\\
Further, we have \\
\begin{equation}
    \|U^\mathcal{A}[q]f \|_2
\le
\prod_{j=1}^{k}
\left(
B_j^\mathcal{A}
(L_j^\mathcal{A})^2
(R_j^\mathcal{A})^2
\right)
\|f\|_2.\label{eq:path-bound}
\end{equation}
Having defined the above operators, we are now in a position to define the feature extractor associated with the network.

\begin{definition}
Let
$
\Omega^\mathcal{A}=
\left\{
\left(
\Psi_k^\mathcal{A},
\mathcal{M}_k^\mathcal{A},
P_k^\mathcal{A}
\right)
:k\in\mathbb{N} \right\}
$
be an $\mathcal{A}$-module sequence. The corresponding feature extractor $
\Phi_{\Omega^\mathcal{A}}^\mathcal{A}:L^2(\mathbb{R}^n)\longrightarrow
\left(L^2(\mathbb{R}^n)\right)^{\mathcal Q}
$
assigns to each signal $f\in L^2(\mathbb{R}^n)$ the feature vector
\[
\Phi_{\Omega^\mathcal{A}}^\mathcal{A}(f)
=
\bigcup_{k=0}^{\infty}
\Phi_{\Omega^\mathcal{A}}^{\,k,\mathcal{A}}(f),
\]
where
\[
\Phi_{\Omega^\mathcal{A}}^{\,k,\mathcal{A}}(f)
=
\left\{
U^\mathcal{A}[q]f
\star_\mathcal{A}
\varphi_k
:
q\in\Lambda^k
\right\},
\]
and $\varphi_k \in $ $
\{g_{\lambda_{k+1}}:\lambda_{k+1}\in\Lambda_{k+1}\}$ denotes the output-generating atom corresponding to the $k$-th layer of the network.
\end{definition}

In order to guarantee the well-posedness of the feature extractor, we impose suitable conditions on the frame bounds and the Lipschitz constants. In fact let $
\Omega^\mathcal{A}=
\left\{
\left(
\Psi_k^\mathcal{A},
\mathcal{M}_k^\mathcal{A},
P_k^\mathcal{A}
\right)
\right\}$
be an $\mathcal{A}$-module sequence. Suppose that $B_k^\mathcal{A}$ is an upper frame bound for the semi-discrete frame $\Psi_k^\mathcal{A}$, and $L_k^\mathcal{A}$ and $R_k^\mathcal{A}$, the Lipschitz constants of the operators $\mathcal{M}_k^\mathcal{A}$ and $P_k^\mathcal{A}$, respectively. Further if\\
\begin{equation}\label{eq:Admissibility}
\max
\left\{
B_k^\mathcal{A},\;
B_k^\mathcal{A}
(L_k^\mathcal{A})^2
(R_k^\mathcal{A})^2
\right\}
\le 1,
\qquad
\forall\,k\in\mathbb N,
\end{equation}
then, the feature extractor
\[
\Phi_{\Omega^\mathcal{A}}^\mathcal{A}:
L^2(\mathbb{R}^n)
\longrightarrow
\left(L^2(\mathbb{R}^n)\right)^{\mathcal Q}
\]
is well defined. In particular, we have the following estimate
    \begin{equation} \label{eq: conv}
        \big| \bigl\|\Phi_{\Omega^\mathcal{A}}^\mathcal{A}(f)\bigr\| \big|^2
        = \sum_{k=0}^\infty \sum_{q \in \Lambda^k} \bigl\|U^\mathcal{A}[q]f \star_\mathcal{A} \varphi_k\bigr\|^2 \leq \|f\|^2.
    \end{equation}
The condition \eqref{eq:Admissibility} is referred to as the \emph{admissibility condition}, and the corresponding module sequence $\Omega^\mathcal{A}$ is called an \emph{admissible $\mathcal{A}$-module sequence}. The primary objective of this section is to investigate the $\mathcal{A}$-translation invariance of the proposed network. Since the network architecture is constructed using the operator $U^\mathcal{A}$, we first examine how this operator behaves under $\mathcal{A}$-translations. The required properties are established below.
\begin{proposition}\label{prop:dilation}
The following identities hold for any $f,g\in L^2(\mathbb{R}^n)$, $t \in \mathbb{R}^n,s \neq 0.$

\begin{align*}
\text{(i)}\quad & 
T_t^\mathcal{A}(f\star_\mathcal{A} g)
=
(T_t^\mathcal{A} f)\star_\mathcal{A} g
=
f\star_\mathcal{A}(T_t^\mathcal{A} g).\\[4pt]
\text{(ii)}\quad &
D_s^\mathcal{A} T_t^\mathcal{A}
=
\exp \bigg(
-\pi i
\sum_{j=1}^{n}
\left(
\frac{t_j^2}{s^2}-t_j^2
\right)
\frac{a_j}{b_j}
\bigg)
T_{t/s}^\mathcal{A}
D_s^\mathcal{A}.
\end{align*}

\end{proposition}
\begin{proof}
    \begin{enumerate}
        \item [(i)] By using \eqref{eq:chirp trans} and \eqref{eq:chirp conv}, we get\\
        
        $\begin{aligned}
            \sqrt{|b_1 \cdot b_2 \cdots b_n}|C_{\frac{a}{b}}(T_t^\mathcal{A}f\star_\mathcal{A} g))(s)
&=C_{\frac{a}{b}}(T_t^\mathcal{A}f) \star C_{\frac{a}{b}}g(s)\\
&= exp \big(\pi i \displaystyle \sum_{j=1}^n \frac{a_j t_j^2}{b_j} \big)T_t(C_{\frac{a}{b}}f \star C_{\frac{a}{b}}g)(s)\\
&= \sqrt{|b_1 \cdot b_2 \cdots b_n}|C_{\frac{a}{b}}(T_t^{\mathcal{A}}(f \star_{\mathcal{A}} g))(s).
        \end{aligned}$\\
        
Hence, we have $T_t^\mathcal{A}(f\star_\mathcal{A} g)
=
(T_t^\mathcal{A} f)\star_\mathcal{A} g.$ Similarly, the other part can be proved. 

\item[(ii)]  By using the fact that $D_sT_t=T_{t/s}D_s$ and \eqref{eq:chirp trans}, we have 

\[
\begin{aligned}
    D_s^\mathcal{A} T_t^\mathcal{A}f(\omega) &= C_{-\frac{a}{b}}D_s C_{\frac{a}{b}}T_t^\mathcal{A}f(\omega)\\
    &= C_{-\frac{a}{b}}D_s \exp \bigg(\pi i \sum_{j=1}^n \frac{a_j t_j^2}{b_j} \bigg) \, T_t\big(C_{\frac{a}{b}} f\big)(\omega) \\
    &=  \exp \bigg(\pi i \sum_{j=1}^n \frac{a_j t_j^2}{b_j} \bigg) \, C_{-\frac{a}{b}} T_{t/s} C_{\frac{a}{b}}C_{-\frac{a}{b}}D_s C_{\frac{a}{b}}f(\omega)\\
    &=  \exp \bigg(\pi i \sum_{j=1}^n \frac{a_j t_j^2}{b_j} \bigg) \, \exp \bigg(-\pi i \sum_{j=1}^n \frac{a_j t_j^2}{ b_j s^2} \bigg) \, T_{t/s}^\mathcal{A} D_s^\mathcal{A}f(\omega)\\
    &= \exp \bigg(-\pi i \sum_{j=1}^{n}\left(\frac{t_j^2}{s^2}-t_j^2\right)\frac{a_j}{b_j}\bigg)\, T_{t/s}^\mathcal{A} D_s^\mathcal{A}f(\omega).
\end{aligned}
\]

    \end{enumerate}
\end{proof}

\begin{figure}[!t]
    \centering
    \begin{subfigure}{0.3\textwidth}
        \centering
        \includegraphics[width=0.8\linewidth]{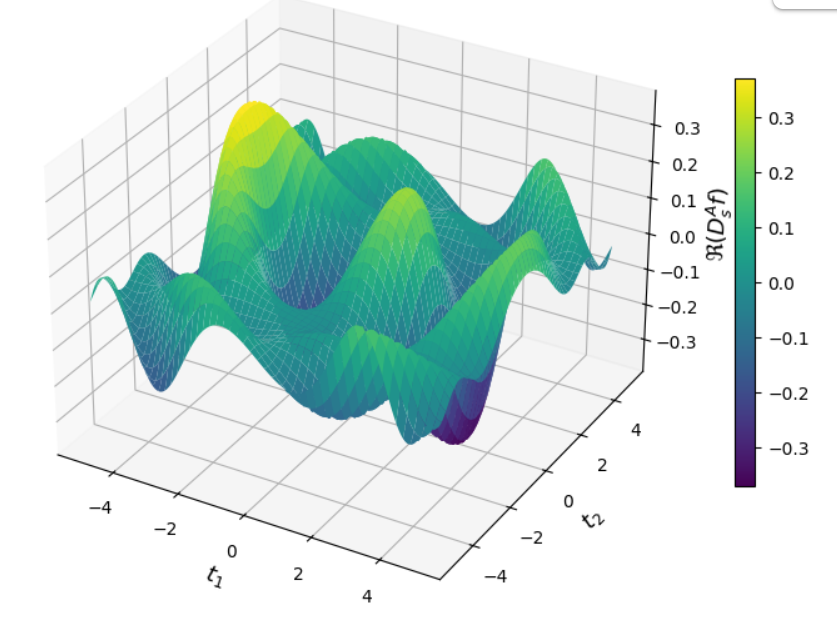}
        \caption{Real part}
        \label{fig:example-5-row1a}
    \end{subfigure}
    \begin{subfigure}{0.3\textwidth}
        \centering
        \includegraphics[width=0.8\linewidth]{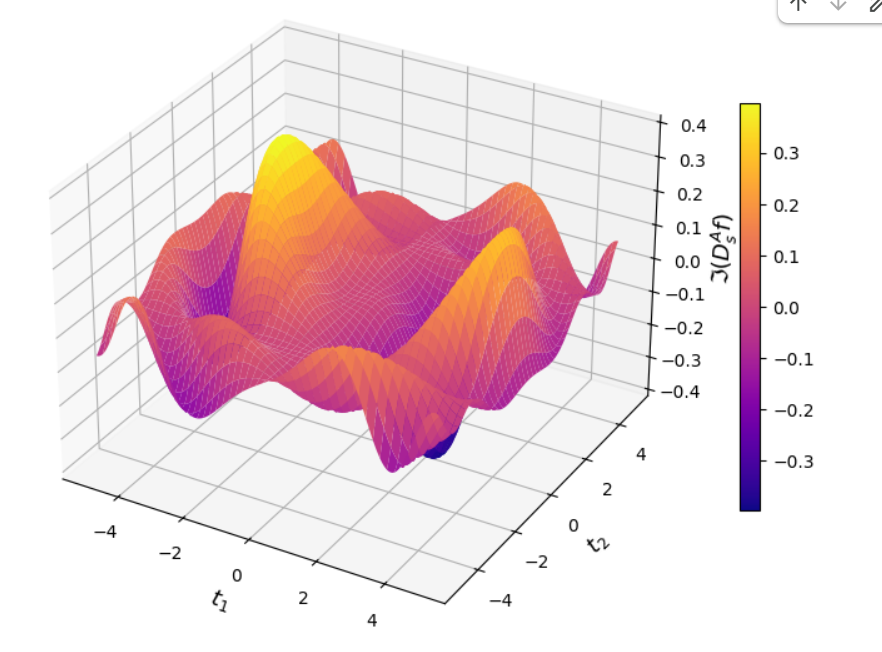}
        \caption{Imaginary part}
        \label{fig:example-5-row1b}
    \end{subfigure}

    \vspace{1em} 

    \caption{The real part and  imaginary part of the $\mathcal{A}$-dilation of $f(t_1,t_2)=\frac{\sin t_1}{t_1}\frac{\sin t_2}{t_2}$ with $a_j=2,b_j=17$ for $j=1,2.$}
    \label{fig:example-5}
\end{figure}

\vspace{0.25cm}

\vspace{0.25cm}
\begin{lemma}
Let $\Omega^{\mathcal{A}}
=\left\{\left(\Psi^{\mathcal{A}}_{k},\,\mathcal{M}^{\mathcal{A}}_{k},\,P^{\mathcal{A}}_{k}\right): k \in \mathbb{N} \right\}$ be an $\mathcal{A}$-module sequence, 
$\left\{g_{\lambda_k}\;:\;\lambda_k\in\Lambda_k\right\}$, the atoms of the frame $\Psi_{\mathcal{A}}^{k}$ and 
$s_k\ge1$, the pooling factor associated with the $k$-th network layer.Assume that, for each $1\le k\le N$, the operator
$T_t^{\mathcal{A}}$ commutes with both
$\mathcal{M}^{\mathcal{A}}_{k}$ and $P^{\mathcal{A}}_{k}$. Then, for every path $q\in\Lambda^N$ and for every $t \in \mathbb{R}^n$, we have\\
\begin{center}

$U^{\mathcal{A}}[q]T_t^{\mathcal{A}}f
=
\exp \bigg(
\pi iN
\displaystyle\sum_{j=1}^{n}\frac{t_j^{2}a_j}{b_j}
\bigg) \exp \bigg(- \pi i(1/s_1^2 + 1/s_2^2 + \cdots 1/s_N^2)\displaystyle\sum_{j=1}^{n}\frac{t_j^{2}a_j}{b_j}\bigg)T_{t/s_1\cdot s_2 \cdots s_N}^{\mathcal{A}}U^{\mathcal{A}}[q]f$.
\end{center}
\end{lemma}

\begin{proof}
    By using Proposition \eqref{prop:dilation}, we have for $\lambda_k \in \Lambda_k$,
    \[
    \begin{aligned}
        U_k^\mathcal{A}[\lambda_k]T_t^{\mathcal{A}}f 
        &=D^{\mathcal{A}}_kP^{\mathcal{A}}_k\mathcal{M}^{\mathcal{A}}_kT_t^{\mathcal{A}}(f \star_\mathcal{A} g_{\lambda_k})\\ 
        &= D^{\mathcal{A}}_kT_t^{\mathcal{A}}P^{\mathcal{A}}_k\mathcal{M}^{\mathcal{A}}_k(f \star_\mathcal{A} g_{\lambda_k})\\
        &= \exp \bigg(\pi i \sum_{j=1}^n \frac{a_j t_j^2}{b_j} \bigg) \exp \bigg(-\pi i \sum_{j=1}^n \frac{a_j t_j^2}{ b_j s_k^2} \bigg)T_{t/s_k}^\mathcal{A}D^{\mathcal{A}}_kP^{\mathcal{A}}_k\mathcal{M}^{\mathcal{A}}_k(f \star_\mathcal{A} g_{\lambda_k})\\
        &=\exp \bigg(\pi i \sum_{j=1}^n \frac{a_j t_j^2}{b_j} \bigg) \exp \bigg(-\pi i \sum_{j=1}^n \frac{a_j t_j^2}{ b_j s_k^2} \bigg)T_{t/s_k}^\mathcal{A} U_k^\mathcal{A}[\lambda_k]f.
    \end{aligned}
    \]

    But,
    $U_k^\mathcal{A}[q]T_t^{\mathcal{A}}f=U_N^\mathcal{A}[\lambda_N] \cdots U_1^\mathcal{A}[\lambda_1]T_t^{\mathcal{A}}f$, from which the result follows.
    
\end{proof}
\vspace{0.5cm}
In the following theorem, we study the asymptotic translation invariance of the proposed feature extractor.\\
\vspace{0.25cm}
\begin{theorem}\label{thm:stability}
   Let $\Omega^{\mathcal{A}}
=\left\{\left(\Psi^{\mathcal{A}}_{k},\,\mathcal{M}^{\mathcal{A}}_{k},\,P^{\mathcal{A}}_{k}\right): k \in \mathbb{N} \right\}$ be an $\mathcal{A}$-module sequence, 
$\left\{g_{\lambda_k}\;:\;\lambda_k\in\Lambda_k\right\}$, the atoms of the frame $\Psi_{\mathcal{A}}^{k}$ and 
$s_k\ge1$, the pooling factor associated with the $k$-th network layer.Assume that the operator
$T_t^{\mathcal{A}}$ commutes with both
$\mathcal{M}^{\mathcal{A}}_{k}$ and $P^{\mathcal{A}}_{k}$. Also, assume that there exists a $K_1>0$ such that all output generating atoms satisfy $|\omega \mathcal{F}_\mathcal{A}(\varphi_k(\omega)|<K_1$ a.e $\omega \in \mathbb{R}^n \quad \text{and for all} \quad k \in \mathbb{N}$. Then, there exists a $K>0$ such that \\
\begin{flalign}\label{eq:invariance}
&\Bigl\|
\exp\bigg(-\pi ik\sum_{j=1}^{n}\frac{t_j^{2}a_j}{b_j}\bigg)
\Phi_{\Omega^\mathcal{A}}^{\,k,\mathcal{A}}(T_t^{\mathcal{A}}f)
-
\Phi_{\Omega^\mathcal{A}}^{\,k,\mathcal{A}}(f)
\Bigr\|&&\notag\\
&\le \Biggl(
\left(\sum_{\ell=1}^{k}\frac{1}{s_\ell^2}\right)^2\sum_{j=1}^{n}\frac{t_j^4a_j^2}{b_j^2}
+\frac{1}{(s_1s_2\cdots s_k)^4}\sum_{j=1}^{n}\frac{t_j^4a_j^2}{b_j^2}
+\frac{4}{(s_1s_2\cdots s_k)^2}\sum_{j=1}^{n}\frac{t_j^2p_j^2}{b_j^2}+\frac{4}{(s_1s_2\cdots s_k)^3}\sum_{j=1}^{n}\frac{|t_j|^3|p_j||a_j|}{b_j^2}
&&\notag\\
&\quad
+\frac{4}{(s_1s_2\cdots s_k)}\left(\sum_{\ell=1}^{k}\frac{1}{s_\ell^2}\right)\sum_{j=1}^{n}\frac{|t_j|^3|p_j||a_j|}{b_j^2}+\frac{8}{(s_1s_2\cdots s_k)^2}\sum_{j=1}^{n}\frac{t_j^2|p_j|}{b_j^2}+\frac{4}{(s_1s_2\cdots s_k)^2}\sum_{j=1}^{n}\frac{t_j^2}{b_j^2}
&&\notag\\
&\quad
+\frac{4}{(s_1s_2\cdots s_k)^3}\sum_{j=1}^{n}\frac{|t_j|^3|a_j|}{b_j^2}+\frac{4}{(s_1s_2\cdots s_k)}\left(\sum_{\ell=1}^{k}\frac{1}{s_\ell^2}\right)\sum_{j=1}^{n}\frac{|t_j|^3|a_j|}{b_j^2}
\Biggr)
K^2\|f\|^2\,n\pi^2.&&
\end{flalign}
\end{theorem}
\begin{proof}
Consider
   \begin{align*}
&\left\|
\exp \bigg(-\pi i k \sum_{j=1}^{n} \frac{ t_j^{2} a_j}{b_j}\bigg)
\Phi_{\Omega^\mathcal{A}}^{k,\mathcal{A}}\!\left(T_t^\mathcal{A}(f)\right)
-
\Phi_{\Omega^\mathcal{A}}^{k,\mathcal{A}}(f)
\right\|^{2} \\[1ex]
&=
\sum_{q\in\Lambda^{k}}
\left\|
\exp \bigg(-\pi i k \sum_{j=1}^{n} \frac{ t_j^{2} a_j}{b_j}\bigg)
\left(U^\mathcal{A}[q]\,T_t^\mathcal{A} f\right)\star_{\mathcal{A}}\varphi_k
-
\left(U^\mathcal{A}[q]\,f\right)\star_{\mathcal{A}}\varphi_k
\right\|^{2} \\[1ex]
&=
\sum_{q\in\Lambda^{k}}
\left\|
\exp \bigg(-\pi i
\left(
\frac{1}{s_1^{2}}+\cdots+\frac{1}{s_k^{2}}
\right)
\sum_{j=1}^{n} \frac{ t_j^{2} a_j}{b_j}\bigg)
\left(
T_{t/(s_1\cdots s_k)}^\mathcal{A}
f_q
\right)
\star_{\mathcal{A}}\varphi_k
-
f_q\star_{\mathcal{A}}\varphi_k
\right\|^{2}.
\end{align*}\\
where $f_q=U^\mathcal{A}[q]f.$ Let $s=s_1\cdots s_k$, $s'=\frac{1}{s_1^{2}}+\cdots+\frac{1}{s_k^{2}} $ and $C_{k,\mathcal{A}}(t)=\exp \bigg(-\pi i
\left(
\frac{1}{s_1^{2}}+\cdots+\frac{1}{s_k^{2}}
\right)
\displaystyle \sum_{j=1}^{n} \frac{ t_j^{2} a_j}{b_j}\bigg).$ Hence, we can write \\
 \begin{equation} \label{eq:simplification}
\left\|
\exp \big(-\pi i k \sum_{j=1}^{n} \frac{ t_j^{2} a_j}{b_j}\bigg)
\Phi_{\Omega^\mathcal{A}}^{k,\mathcal{A}}\!\left(T_t^\mathcal{A}(f)\right)
-
\Phi_{\Omega^\mathcal{A}}^{k,\mathcal{A}}(f)
\right\|^{2}
= \sum_{q\in\Lambda^{k}} \left\| C_{k,\mathcal{A}}(t) \left(
T_{t/(s_1\cdots s_k)}^\mathcal{A}
f_q
\right)
\star_{\mathcal{A}}\varphi_k
-
f_q\star_{\mathcal{A}}\varphi_k
\right\|^{2}.
\end{equation}\\
By applying the Parseval formula for SAFT , we can write the LHS of \eqref{eq:simplification} \\
\begin{align*}
      &= \sum_{q\in\Lambda^{k}} \int_{\mathbb{R}^n}\big|C_{k,\mathcal{A}}(t) \left(
\mathcal{F}_\mathcal{A}((T_{t/(s_1\cdots s_k)}^\mathcal{A}
f_q
\right)
\star_{\mathcal{A}}\varphi_k)(\omega)
-
\mathcal{F}_\mathcal{A}(f_q\star_{\mathcal{A}}\varphi_k)(\omega)\big |
^{2}d \omega \\
&= \sum_{q\in\Lambda^{k}} \int_{\mathbb{R}^n}\big|C_{k,\mathcal{A}}(t) \left(
\mathcal{F}_\mathcal{A}((T_{t/(s_1\cdots s_k)}^\mathcal{A}
f_q
\right)(\omega)(\mathcal{F}_\mathcal{A}(
\varphi_k)(\omega)
-
\mathcal{F}_\mathcal{A}(f_q(\omega)\mathcal{F}_\mathcal{A}(\varphi_k)(\omega)\big |
^{2}d \omega.\\
\end{align*}\\
But by \eqref{eq:mod trans}, $\mathcal{F_A}(T_{t/s}^\mathcal{A}f)(\omega)= exp \big(\frac{\pi i}{s^2} \displaystyle \sum_{j=1}^n \frac{a_j t_j^2}{b_j}+ \frac{2 \pi i}{s}  \displaystyle \sum_{j=1}^n \frac{p_jt_j-t_jw_j}{b_j} \bigg)\mathcal{F_\mathcal{A}}(f)(\omega).$
  
Thus, we get the LHS of \eqref{eq:simplification}\\
\begin{align*}
    &= \sum_{q\in\Lambda^{k}} \int_{\mathbb{R}^n}\bigg|C_{k,\mathcal{A}}(t)\exp \big(\frac{\pi i}{s^2} \displaystyle \sum_{j=1}^n \frac{a_j t_j^2}{b_j}+ \frac{2 \pi i}{s}  \displaystyle \sum_{j=1}^n \frac{2p_jt_j-2t_jw_j}{b_j} \bigg)-1 \bigg|^2 \bigg|\mathcal{F}_\mathcal{A}(f_q(\omega)\mathcal{F}_\mathcal{A}(\varphi_k)(\omega)\bigg |^2 d\omega.\\
\end{align*}
For simplicity, we use some notations. \\
Let $\alpha_j= s'\frac{ t_j^{2} a_j}{b_j},
    \beta_j=\frac{a_j t_j^2}{s^2b_j}$ and $\gamma_j= \frac{2p_jt_j-2t_jw_j}{sb_j}.$ Write $\xi_j= - s'\frac{ t_j^{2} a_j}{b_j}+   \frac{a_j t_j^2}{s^2b_j}+    \frac{2p_jt_j-2t_jw_j}{sb_j}  $. Then, we have\\
    \begin{equation}\label{eq:triangle}
\begin{aligned}
\xi_j^2
&=
\alpha_j^2+\beta_j^2+\gamma_j^2
-2\alpha_j\beta_j
+\beta_j\gamma_j
-2\gamma_j\alpha_j \\
&\leq
\alpha_j^2+\beta_j^2+\gamma_j^2
+2\beta_j\gamma_j
-2\gamma_j\alpha_j \\
&\leq
\alpha_j^2+\beta_j^2+\gamma_j^2
+2|\beta_j\gamma_j|
+2|\gamma_j\alpha_j|.
\end{aligned}
    \end{equation}
But
\begin{equation}\label{eq:3}
\begin{aligned}
\gamma_j^2 &\leq \underbrace{\frac{4t_j^2 p_j^2}{s^2b_j^2}}_{:=\gamma_{j,1}}
+ \underbrace{\frac{8t_j^2|p_j|}{s^2 b_j^2}}_{:=\gamma_{j,2}}| \omega_j|
+ \underbrace{\frac{4t_j^2 }{s^2b_j^2}}_{:=\gamma_{j,3}}\omega_j^2, \\
|\beta_j \gamma_j| &\leq \underbrace{\frac{2|a_j||t_j|^3|p_j|}{s^3 b_j^2}}_{:=\theta_{j,1}}
+ \underbrace{\frac{2|a_j||t_j|^3}{s^3 b_j^2}}_{:=\theta_{j,2}}|\omega_j|, \\
|\alpha_j \gamma_j| &\leq \underbrace{\frac{2|a_j|s'|t_j|^3|p_j|}{sb_j^2}}_{:=\lambda_{j,1}}
+ \underbrace{\frac{2|a_j|s'|t_j|^3}{s b_j^2}}_{:=\lambda_{j,2}}|\omega_j|.
\end{aligned}
\end{equation}
    \vspace{0.5cm}
    Write $ B_{k,A_j}(t_j,\omega_j)=\exp \big( \pi i \xi_j\big)$. Then,\\
    \begin{equation}
\left|B_{k,A_j}(t_j,\omega_j)-1\right|^2
=
\left|\exp \big( \pi i \xi_j\big)-1\right|^2
=
4\sin^2\left(\frac{\pi\xi_j}{2}\right)
\leq
\pi^2\xi_j^2.
\end{equation}
    \vspace{0.25cm}
    Define
\begin{center}
  $
  D_{k,\mathcal{A}}(t, \omega)
  =B_{k,A_1}(t_1,\omega_1) \cdot B_{k,A_2}(t_2,\omega_2)\cdots B_{k,A_n}(t_n,\omega_n).
  $
 \end{center}
 Then, \\
 \begin{center}
     $ D_{k,\mathcal{A}}(t, \omega)=C_{k,\mathcal{A}}(t)
     \exp \big(\frac{\pi i}{s^2} \displaystyle \sum_{j=1}^n \frac{a_j t_j^2}{b_j}+ \frac{2 \pi i}{s}  \displaystyle \sum_{j=1}^n \frac{p_jt_j-t_jw_j}{b_j} \bigg).      $
 \end{center}

Further, it can be easily shown that \\
\begin{equation} \label{eq:DB}
\begin{aligned}
   \big|  D_{k,\mathcal{A}}(t, \omega)-1|^2 & \leq n \displaystyle \sum_{j=1}^n\big | B_{k,A_j}(t_j,\omega_j)-1\big|^2.\\
\end{aligned}\\
\end{equation}
\vspace{0.25cm}
Thus the LHS of \eqref{eq:simplification} can be written as\\
\begin{align*}
    &= \sum_{q\in\Lambda^{k}} \int_{\mathbb{R}^n}\bigg|D_{k,\mathcal{A}}(t,\omega)-1 \bigg|^2 \bigg|\mathcal{F}_\mathcal{A}(f_q(\omega)\mathcal{F}_\mathcal{A}(\varphi_k)(\omega)\bigg |^2 d\omega.\\
\end{align*}
By using \eqref{eq:triangle}, \eqref{eq:3}, \eqref{eq:DB}, we have,\\
\begin{flushleft}
\begin{equation}
\begin{aligned}
\frac{1}{n \pi^2}&\Biggl\|
\exp \bigg(-\pi ik\displaystyle\sum_{j=1}^{k}\frac{t_j^{2}a_j}{b_j}\bigg)
\Phi_{\Omega^\mathcal{A}}^{\,k,\mathcal{A}}(T_t^{\mathcal{A}}f)
-
\Phi_{\Omega^\mathcal{A}}^{\,k,\mathcal{A}}(f)
\Biggr\| \\[1ex]
&\quad\leq
\displaystyle\sum_{j=1}^n \alpha_j^2
\sum_{q\in\Lambda^k}
\int_{\mathbb{R}^n}
\Bigl|
\mathcal{F}_{\mathcal{A}}
\bigl(f_q(\omega)\mathcal{F}_{\mathcal{A}}(\varphi_k)(\omega)\bigr)
\Bigr|^2\,d\omega +
\displaystyle\sum_{j=1}^n \beta_j^2
\sum_{q\in\Lambda^k}
\int_{\mathbb{R}^n}
\Bigl|
\mathcal{F}_{\mathcal{A}}
\bigl(f_q(\omega)\mathcal{F}_{\mathcal{A}}(\varphi_k)(\omega)\bigr)
\Bigr|^2\,d\omega \\[1ex]
&\qquad+ \displaystyle\sum_{j=1}^n \gamma_{j,1}
\sum_{q\in\Lambda^k}
\int_{\mathbb{R}^n}
\Bigl|
\mathcal{F}_{\mathcal{A}}
\bigl(f_q(\omega)\mathcal{F}_{\mathcal{A}}(\varphi_k)(\omega)\bigr)\Bigl|^2 d\omega
\\[1ex]
&\qquad  +
\displaystyle\sum_{j=1}^n \gamma_{j,2}
\sum_{q\in\Lambda^k}
\int_{\mathbb{R}^n}
\Bigl|
 \omega_j \mathcal{F}_{\mathcal{A}}
\bigl(f_q(\omega)\mathcal{F}_{\mathcal{A}}(\varphi_k)(\omega)\bigr)
\Bigr|^2\,d\omega+ \displaystyle\sum_{j=1}^n \gamma_{j,3}
\sum_{q\in\Lambda^k}
\int_{\mathbb{R}^n}
\Bigl|
 \omega_j^2 \mathcal{F}_{\mathcal{A}}
\bigl(f_q(\omega)\mathcal{F}_{\mathcal{A}}(\varphi_k)(\omega)\bigr)
\Bigr|^2\,d\omega
\\[1ex]
&\qquad
+\displaystyle\sum_{j=1}^n 2 \theta_{j,1}
\sum_{q\in\Lambda^k}
\int_{\mathbb{R}^n}
\Bigl|
\mathcal{F}_{\mathcal{A}}
\bigl(f_q(\omega)\mathcal{F}_{\mathcal{A}}(\varphi_k)(\omega)\bigr)
\Bigr|^2\,d\omega+
\displaystyle\sum_{j=1}^n 2 \theta_{j,2}
\sum_{q\in\Lambda^k}
\int_{\mathbb{R}^n}
\Bigl|
 \omega_j \mathcal{F}_{\mathcal{A}}
\bigl(f_q(\omega)\mathcal{F}_{\mathcal{A}}(\varphi_k)(\omega)\bigr)
\Bigr|^2\,d\omega 
\\[1ex]
&\qquad
+\displaystyle\sum_{j=1}^n 2 \lambda_{j,1}
\sum_{q\in\Lambda^k}
\int_{\mathbb{R}^n}
\Bigl|
\mathcal{F}_{\mathcal{A}}
\bigl(f_q(\omega)\mathcal{F}_{\mathcal{A}}(\varphi_k)(\omega)\bigr)
\Bigr|^2\,d\omega+
\displaystyle\sum_{j=1}^n 2 \lambda_{j,2}
\sum_{q\in\Lambda^k}
\int_{\mathbb{R}^n}
\Bigl|
 \omega_j \mathcal{F}_{\mathcal{A}}
\bigl(f_q(\omega)\mathcal{F}_{\mathcal{A}}(\varphi_k)(\omega)\bigr)
\Bigr|^2\,d\omega. 
\end{aligned}\label{eq:final}
\end{equation}
\end{flushleft}
Then, using Riemann Lebesgue lemma for SAFT, (see \cite{MR4458766}), $|\mathcal{F}_\mathcal{A}(\varphi_k)| \leq K_2$ for some $K_2 >0.$ On the other hand,\\
\begin{equation}
\begin{aligned}
  |\omega_j|^{1/2} |\mathcal{F}_{\mathcal{A}}(\varphi_k)(\omega)| \leq |\omega|^{1/2} |\mathcal{F}_{\mathcal{A}}(\varphi_k)(\omega)| \leq  \begin{cases}
      \frac{K_1}{|\omega|^1/2} \leq K_1, & | \omega| >1,\\
      K_2, & | \omega| \leq 1
  \end{cases}.
\end{aligned}
\end{equation}\\
Let $K:=$max\{$K_1,K_2$\}. Then $|\omega_j|^{1/2} |\mathcal{F}_{\mathcal{A}}(\varphi_k)(\omega)| \leq K$. But we already have $|\omega_j| |\mathcal{F}_{\mathcal{A}}(\varphi_k)(\omega)| \leq K$, $ |\mathcal{F}_{\mathcal{A}}(\varphi_k)(\omega)| \leq K$. But by  \eqref{eq:path-bound}, $\sum_{q\in\Lambda^k}
\int_{\mathbb{R}^n}
\Bigl|
\mathcal{F}_{\mathcal{A}}
\bigl(f_q(\omega) \Bigl|^2d \omega \leq ||f||^2$. Thus, on substitution of these estimates in \eqref{eq:final}, we get ~\eqref{eq:invariance}, proving our assertion. 

\end{proof}

\vspace{0.25cm}
\begin{corollary}
    For an appropriate choice of pooling factors \{$s_k$\}, the right-hand side of ~\eqref{eq:invariance} can be made arbitrarily small.
\end{corollary}
\begin{proof}
The proof is similar to Corollary 1 of \cite{MR4983277}, hence it is omitted.
\end{proof}

In \cite{MR2957703}, Mallat established the translation invariance of his wavelet-based network. In contrast, the network considered here is not necessarily translation-invariant in general. The following corollary shows that, for a suitable choice of pooling factors, the network associated with the SAFT is asymptotically invariant under the $\mathcal{A}$-translation.

\begin{corollary}
In addition to the hypotheses of Theorem~\ref{thm:stability}, if we assume that there exists a $K_2>0$ such that $|\mathcal{F}_\mathcal{A}(\varphi_k)| \leq K_2$ for all $k \in \mathbb{N}$, then the network associated with the SAFT is asymptotically $\mathcal{A}$-translation invariant for a suitable choice of pooling factors.
\end{corollary}
\begin{proof}
The proof is again similar to Corollary 2 of \cite{MR4983277}, hence it is omitted.
\end{proof}
\begin{corollary}
  Let $\Omega^{\mathcal{A}}
=\left\{\left(\Psi^{\mathcal{A}}_{k},\,\mathcal{M}^{\mathcal{A}}_{k},\,P^{\mathcal{A}}_{k}\right): k \in \mathbb{N} \right\}$ be an $\mathcal{A}$-module sequence, 
$\left\{g_{\lambda_k}\;:\;\lambda_k\in\Lambda_k\right\}$, the atoms of the frame $\Psi_{\mathcal{A}}^{k}$ and 
$s_k\ge1$, the pooling factor associated with the $k$-th network layer. Assume that the operator
$T_t^{\mathcal{A}}$ commutes with both
$\mathcal{M}^{\mathcal{A}}_{k}$ and $P^{\mathcal{A}}_{k}$. Also, assume that there exists a $K_1>0$ such that all output generating atoms satisfy $|\omega \mathcal{F}_\mathcal{A}(\varphi_k(\omega)|<K_1$ a.e $\omega \in \mathbb{R}^n$ and for all $k \in \mathbb{N}$. Then, there exists a $K>0$ such that \\
\begin{flalign}
&\left\| 
\exp\left(-\pi ik\sum_{j=1}^{n}\frac{t_j^{2}a_j}{b_j}\right)
\Phi_{\Omega^\mathcal{A}}^{\,k,\mathcal{A}}(T_t^{\mathcal{A}}f)
-
\Phi_{\Omega^\mathcal{A}}^{\,k,\mathcal{A}}(f)
\right\|^2 &&\notag\\
&\leq \Bigg[
\left( \frac{1}{s_1^2} + \frac{1}{s_2^2} + \cdots + \frac{1}{s_k^2} \right)^2
\sum_{j=1}^{n}\frac{t_j^4a_j^2}{b_j^2}
+
\left(1-\frac{1}{(s_1s_2\cdots s_k)^2}\right)^2
\sum_{j=1}^{n}\frac{t_j^4a_j^2}{b_j^2} &&\notag\\
&\quad + 2\left( \frac{1}{s_1^2} + \frac{1}{s_2^2} + \cdots + \frac{1}{s_k^2} \right)
\left(1-\frac{1}{(s_1s_2\cdots s_k)^2}\right)
\sum_{j=1}^{n}\frac{t_j^4a_j^2}{b_j^2} + 4\left( 1- \frac{1}{s_1s_2\cdots s_k}\right)^2
\sum_{j=1}^{n}\frac{t_j^2p_j^2}{b_j^2}&&\notag \\
&\quad + 4\left( 1- \frac{1}{(s_1s_2\cdots s_k)^2}\right)
\left( 1- \frac{1}{s_1s_2\cdots s_k}\right)
\sum_{j=1}^{n}\frac{|t_j|^3|p_j||a_j|}{b_j^2}&&\notag\\
&\quad+ 4\left( 1- \frac{1}{(s_1s_2\cdots s_k)^2}\right)
\left( 1- \frac{1}{s_1s_2\cdots s_k}\right)
\sum_{j=1}^{n}\frac{|t_j|^3|a_j|}{b_j^2}&&\notag \\
&\quad + 8\left( 1- \frac{1}{s_1s_2\cdots s_k}\right)^2
\sum_{j=1}^{n}\frac{t_j^2|p_j|}{b_j^2}
+
4\left( 1- \frac{1}{s_1s_2\cdots s_k}\right)^2
\sum_{j=1}^{n}\frac{t_j^2}{b_j^2}&&\notag \\
&\quad+ 4\left( \frac{1}{s_1^2} + \frac{1}{s_2^2} + \cdots + \frac{1}{s_k^2} \right)
\left( 1- \frac{1}{s_1s_2\cdots s_k}\right)
\sum_{j=1}^{n}\frac{|t_j|^3|a_j|}{b_j^2} &&\notag\\
&\quad + 4\left( 1- \frac{1}{s_1s_2\cdots s_k}\right)
\left( \frac{1}{s_1^2} + \frac{1}{s_2^2} + \cdots + \frac{1}{s_k^2} \right)
\sum_{j=1}^{n}\frac{|t_j|^3|a_j|}{b_j^2}
\Bigg] K^2\|f\|^2 n\pi^2. &&
\end{flalign}

\end{corollary}
\begin{proof}
    The proof is similar to that of Theorem~\ref{thm:stability}, so it is left to the reader. 
\end{proof}
\vspace{0.5cm}
\begin{example}
    Choose a mother wavelet constructed using multiresolution analysis (MRA), say Haar wavelet $\psi$. (See Daubechies \cite{daubechies1992ten} for the study of multiresolution analysis and the construction of wavelets.) Define $\phi_k(x)= C_{- a}2^k \psi(2^kx).$ Consider SAFT of $\phi_k$ with $b=1$ and $p=0$. Then, using Proposition~\ref{prop:equivalence}, we can show that the collection $\Psi=$\{ $T_s^\mathcal{A}\check{\phi_k},  s \in \mathbb{R}, k \in \mathbb{N} $ \} forms a semi-discrete frame for $L^2(\mathbb{R})$. We employ the semi-discrete frame $\Psi$ at each layer of the network, with
$\mathcal{C}_{-a}\psi$ serving as the output-generating atom at each level.
Let $\mathcal{M}^{\mathcal{A}}:=C_{-a}TC_{a}$ denote the
nonlinearity at each layer, where $Tf(x)=|f(x)|$. Let
$P^{\mathcal{A}}:=\operatorname{Id}$ denote the identity operator on
$L^{2}(\mathbb{R})$, which acts as the pooling operator in each layer of the
network. Further, we choose $k^{2}$ as the scaling factor at the
$k$-th level of the network. With these choices, we obtain the following
error estimate for $f_1=C_{-a}\chi_{[3,5]}+2  C_{-a}\chi_{[1,2]}+ 3  C_{-a}\chi_{[-0.5,0.5]}$, $ f_2=C_{-a}\chi_{[10,12]}+4 C_{-a}\chi_{[4,6]}+ 9  C_{-a}\chi_{[0,1]}$ and for different values of $ q$ and $t$ using a Python-based implementation:
\begin{equation}
E[q](f,t):=
\left\|
\left(U^{\mathcal{A}}[q]T_{t}^{\mathcal{A}}f\right)\star_{\mathcal{A}}\mathcal{C}_{-a}\psi
-
\left(U^{\mathcal{A}}[q]f\right)\star_{\mathcal{A}}C_{-a}\psi
\right\|.
\end{equation}
The results are displayed in Table \eqref{tab:1d}. \label{ex:1D}
\end{example}

\begin{example}
    Similar to the 1-D case, we employ the same technique for a 2-D semi-discrete frame. Choose a mother wavelet $\psi$ and father wavelet $\phi$ constructed using MRA. As before, we choose the Haar wavelet. Define

\[
\begin{aligned}
\psi^{(0,0)} := \phi \otimes \phi,\quad
\psi^{(1,0)} := \psi \otimes \phi,\quad
\psi^{(0,1)} := \phi \otimes \psi,\quad
\psi^{(1,1)} := \psi \otimes \psi.
\end{aligned}
\]

Define $g_{e,k}(x)= C_{-a} 2^{2k} \psi^e(2^kx)$ where $e \in \{(0,0),(0,1),(1,0),(1,1)\}$. Consider SAFT of $g_{e,k}$ with $b_j=1$ and $p_j=0$ for $j=1,2$. Then, from Proposition \eqref{prop:equivalence}, we can show that $\Psi=$\{ $T_s^\mathcal{A}\check{g_{e,k}}, \quad s \in \mathbb{R}, k \in \mathbb{N} $, $e \in \{(0,0),(0,1),(1,0),(1,1)\}\} $ forms a semi discrete frame for $L^2(\mathbb{R}^2)$.\\

As in the previous example, we choose the same non-linearity, pooling function and the scaling factor with $C_{-a} \psi^{0,0}$ being the output generating function at each level. With these choices, we obtain the following error estimates for $f_3=C_{-a}(\chi_{[3,5]} \otimes \chi_{[3,5]}) +2  C_{-a}(\chi_{[1,2]}\otimes \chi_{[1,2]})+ 3  C_{-a}(\chi_{[-0.5,0.5]}\otimes \chi_{[-0.5,0.5]})$, $ f_4=C_{-a}(\chi_{[10,12]} \otimes \chi_{[10,12]})+4 C_{-a}(\chi_{[4,6]}\otimes \chi_{[4,6]})+ 9  C_{-a}(\chi_{[0,1]}\otimes \chi_{[0,1]})$ and for different values of $ q$ and $t$ using a Python-based implementation:
\begin{equation}
E[q](f,t):=
\left\|
\left(U^{\mathcal{A}}[q]T_{t}^{\mathcal{A}}f\right)\star_{\mathcal{A}}\mathcal{C}_{-a}\psi
-
\left(U^{\mathcal{A}}[q]f\right)\star_{\mathcal{A}}C_{-a}\psi
\right\|.
\end{equation}
The results are displayed in Table \eqref{tab:2d}.\label{ex:2D}
\end{example}
\vspace{0.25cm}
\begin{remark}
   In the one-dimensional case, the error becomes approximately zero by the third layer, whereas in the two-dimensional case, it vanishes by the second layer. This observation is consistent with Theorem~\ref{thm:stability}.

\end{remark}
\vspace{0.75cm}
\section{\textbf{Deformation insensitivity analysis }}
\vspace{0.5cm}
We denote $C^k(\mathbb{R}^n,\mathbb{R}^n)$ as the space of $k$ times continuously differentiable vector fields. More precisely, for a vector field $\tau:\mathbb{R}^n \rightarrow\mathbb{R}^n$, $D \tau$ denotes its Jacobian matrix and $D^k\tau=D(D^{k-1}\tau)$ with the corresponding norm $||D^k\tau||_\infty=\sup_{x \in \mathbb{R}^n}|D^k\tau(x)|_\infty.$\\

In this section, we establish bounds on the insensitivity of the feature extractor $\Phi_{\Omega^\mathcal{A}}^\mathcal{A}$ with respect to deformations of the form $(F_{\tau}f)(x):=f(x-\tau(x)).$\\

The deformation insensitivity bounds obtained below are specific to the signal class under consideration, in the sense that they apply to input signals belonging to a prescribed class. In particular, the signal class is required to possess an inherent insensitivity to deformations, as formalized in the following definition.

\begin{definition}
A signal class $\mathcal{C}\subseteq L^{2}(\mathbb{R}^{n})$ is said to be \emph{deformation-insensitive} if there exist constants $\alpha,C>0$ such that, for every $f\in\mathcal{C}$ and every (possibly nonlinear) $\tau\in C^{1}(\mathbb{R}^{n},\mathbb{R}^{n})$ satisfying
$
||\tau||_{\infty}<\frac{1}{2}
\quad\text{and}\quad
||D\tau||_{\infty}\leq\frac{1}{2n},
$
the following inequality holds:
\begin{equation}
||f-F_{\tau}f||_{2}
\leq
C||\tau||_{\infty}^{\alpha}||f||_{2}.
\end{equation}
\end{definition}
The constants $C>0$ and $\alpha>0$ in the definition depend on the particular signal class $\mathcal{C}$. The parameter $\alpha$ determines the rate at which the deformation error $||f-F_{\tau}f||_{2}$
decreases as the magnitude of the deformation $||\tau||_{\infty}$ tends to zero. In particular, a larger value of $\alpha$ corresponds to a faster decay of the deformation error for increasingly small deformations.
\vspace{0.5cm}
\begin{theorem}\label{thm:feature d}
Let
$
\Omega^\mathcal{A}=$\{($\Psi_n^\mathcal{A},\mathcal{M}_n^\mathcal{A},P_n^\mathcal{A}):n \in \mathbb{N}$\}$
$
be an admissible $\mathcal{A}$-module-sequence, and $\mathcal{C}\subseteq L^{2}(\mathbb{R}^{n})$ be a deformation-insensitive signal class. Then, there exist constants $\alpha,C>0$, independent of $\Omega^\mathcal{A}$, such that for every $f\in\mathcal{C}$ and every $\tau\in C^{1}(\mathbb{R}^{n},\mathbb{R}^{n})$ satisfying
$
||\tau||_{\infty}<\frac{1}{2}$
and $
||D\tau||_{\infty}\leq\frac{1}{2n},
$
the feature extractor $\Phi_{\Omega^\mathcal{A}}^\mathcal{A}$ satisfies
\begin{equation}
\big| \bigl \|
\Phi_{\Omega^\mathcal{A}}^\mathcal{A}(F_{\tau}f)-\Phi^\mathcal{A}_{\Omega^\mathcal{A}}(f)
\bigr \| \big|
\leq
C||\tau||_{\infty}^{\alpha}||f||_{2}.
\end{equation}
\end{theorem}
\begin{proof}
    The proof is similar to that of Proposition 4 of Appendix I of \cite{MR3766318}, and hence is left to the reader. 
\end{proof}
\vspace{0.5cm}
\subsection{\textbf{Bounds for \texorpdfstring{$\mathcal{A}$}{A}-bandlimited functions}}\mbox{}\\

Let $B_L(0)$ denote the unit ball of radius $L$ centered at $0$ in $\mathbb{R}^n.$ Let $L_L^{2,\mathcal{A}}(\mathbb{R}^n):=$\{ $f \in L^2(\mathbb{R}^n):$ supp($\mathcal{F}_\mathcal{A})(f) \subseteq B_L(0)$\} denote the class of $\mathcal{A}$-bandlimited functions. Our aim is to investigate the insensitivity to deformation of functions in  $L_L^{2,\mathcal{A}}(\mathbb{R}^n)$.\\
In order to state the result in this direction, we make use of the following lemmas.\\
\begin{lemma}(\cite{MR3766318})\label{lem:k}
    Let $\gamma \in S(\mathbb{R}^{n})$, the Schwartz class of rapidly decreasing functions in $\mathbb{R}^{n}$ and $\tau\in C^{1}(\mathbb{R}^{n},\mathbb{R}^{n})$ satisfy $||D\tau||_{\infty}\leq\frac{1}{2n},
$. Define $k(x,u)= \gamma(x-\tau(x)-u)- \gamma(x-u).$ Then\\
\begin{center}
    $$ \displaystyle \sup_{x \in \mathbb{R}^n}\int_{\mathbb{R}^n}| k(x,u) | du \leq || \tau||_\infty || \nabla \gamma ||_1,$$
\end{center}
where $\nabla \gamma$ denotes the gradient of $\gamma.$
\end{lemma}
\begin{lemma}(\cite{folland1999real})\label{lem:cor}
If $f$ is a measurable function on $\mathbb{R}^n,$ non-negative or integrabe such that $f(x)=g(||x||_\infty)$ for some function $g$ on $(0,\infty),$ then\\
\begin{center}
    $$ \int_{\mathbb{R}^n}f(x) dx= \sigma(S^{n-1}) \int_{(0,\infty)}g(r)r^{n-1}dr,$$
\end{center}
where $\sigma$ is a measure on $S^{n-1}$.
\end{lemma}
\begin{proposition}\label{prop:band}
 There exists a constant $C>0$, such that for every $f\in L_L^{2,\mathcal{A}}(\mathbb{R}^n)$ and every $\tau\in C^{1}(\mathbb{R}^{n},\mathbb{R}^{n})$ satisfying
$
||D\tau||_{\infty}\leq\frac{1}{2n},
$ it holds that\\
\begin{equation*}
    ||M_g F_\tau f -f||_2 \leq C ||\tau||_\infty||f||_2,
\end{equation*}
where $M_g$ is the multiplication operator with $g(x)=\exp\big(-2 \pi i \displaystyle \sum_{j=1}^n \frac{x_j \tau(x)_j a_j}{b_j}\big). $
\end{proposition}
\begin{proof}
Choose a $\gamma \in S(\mathbb{R}^{n})$ such that $\mathcal{F}_\mathcal{A}(\gamma)= \eta_\mathcal{A}$ on $B_L(0)$.  Then, for all $f \in L_L^{2,\mathcal{A}}(\mathbb{R}^n),
    \mathcal{F}_\mathcal{A}(f \star_\mathcal{A} \gamma)(\omega)= \overline{\eta_\mathcal{A}(\omega)}\mathcal{F}_\mathcal{A}(f)(\omega)\mathcal{F}_\mathcal{A}(\gamma)(\omega)= \mathcal{F}_\mathcal{A}(f)(\omega)$. Hence, $f \star_\mathcal{A} \gamma =f$.\\
    Define, for $f\in L^{2}(\mathbb{R}^n)$, $A_{\gamma}f=f \star_\mathcal{A} \gamma$. Then, $A_{\gamma}f=f$ on $L_L^{2,\mathcal{A}}(\mathbb{R}^d)$. Now, define $g(x)=\exp \big(-2 \pi i \displaystyle \sum_{j=1}^n \frac{x_j \tau(x)_j a_j}{b_j}\big)$. Since $g \in L^\infty(\mathbb{R}^{n})$, $M_g \in B(L^2(\mathbb{R}^{n}))$. Further, for $f\in L^{2}(\mathbb{R}^n)$, define $Kf=(M_gF_\tau A_\gamma - A_\gamma)f.$ Then\\
\begin{equation}\label{eq:Kf}
    Kf=M_gF_\tau f - f, \quad f \in L_L^{2,\mathcal{A}}(\mathbb{R}^n).
\end{equation}
\begin{flushleft}
Consider \\
\end{flushleft}
\begin{align*}
    Kf(x) 
    &= g(x)(A_\gamma f(x-\tau(x))-A_\gamma f(x).\\
    &=g(x)f \star \gamma(x-\tau(x))-f \star \gamma(x)\\
    &=\int_{\mathbb{R}^n} (g(x) T_u^\mathcal{A} \gamma(x- \tau(x)) - T_u^\mathcal{A} \gamma(x))f(u)du.\\ 
\end{align*}\\
Let $k(x,u)= g(x) T_u^\mathcal{A} \gamma(x- \tau(x)) - T_u^\mathcal{A} \gamma(x).$ Then $k$ is clearly locally integrable.  But,
\\
\vspace{0.25cm}
\begin{align*}
k(x,u)&= \exp \big(-2 \pi i \displaystyle \sum_{j=1}^n \frac{x_j \tau(x)_j a_j}{b_j}\big)\exp\bigg( -2 \pi i \displaystyle \sum_{j=1}^n \frac{u_j (x_j-\tau(x)_j-u_j) a_j}{b_j}\bigg) \gamma(x-\tau(x)-u)\\
& -  \exp \bigg( -2 \pi i \displaystyle \sum_{j=1}^n \frac{u_j (x_j-u_j) a_j}{b_j}\bigg) \gamma(x-u).
\end{align*}\\
\begin{table}[!t]
\centering
\caption{Error Estimates for Example \eqref{ex:1D}}
\label{tab:1d}
{\small
\begin{tabular}{|ccccc|}
\hline
Function & Translation & Layer 1 & Layer 2 & Layer 3 \\
& parameter, $t$ & $q=[0]$ & $q=[0,1]$ & $q=[0,1,2]$ \\
\hline
$f_1$
& $20$ & $0.5034$ & $0.002659$ & $0$ \\[1mm]
$f_2$
& $-5$ & $1.7515$ & $0.005964$ & $0$ \\[1mm]
\hline
\end{tabular}
}
\end{table}
\vspace{0.25cm}

\begin{table}[!t]
\centering
\caption{Error Estimates for Example \eqref{ex:2D}}
\label{tab:2d}
{\small
\begin{tabular}{|cccc|}
\hline
Function & Translation & Layer 1 & Layer 2 \\
& parameter, $t$ & $q=[0]$ & $q=[0,1]$ \\
\hline
$f_3$ & $20$ & $0.3280$ & $0$ \\[1mm]
$f_4$ & $-5$ & $0.9151$ & $0$ \\[1mm]
\hline
\end{tabular}
}
\end{table}

Now,
\begin{align*}
    |k(x,u)|
    & \leq \bigg| \exp \bigg( -2 \pi i \displaystyle \sum_{j=1}^n \frac{\tau(x)_j (x_j-u_j) a_j}{b_j}\bigg)(\gamma(x-\tau(x)-u)- \gamma(x-u) )\bigg|\\
    & + \bigg|\bigg(  \exp \bigg( -2 \pi i \displaystyle \sum_{j=1}^n \frac{\tau(x)_j (x_j-u_j) a_j}{b_j}\bigg)-1 \bigg) \gamma(x-u) \bigg|\\
    & = \bigg| \gamma(x-\tau(x)-u)- \gamma(x-u)\bigg| +  \bigg|\bigg(  \exp \bigg( -2 \pi i \displaystyle \sum_{j=1}^n \frac{\tau(x)_j (x_j-u_j) a_j}{b_j}\bigg)-1 \bigg) \gamma(x-u) \bigg|\\
    & \leq \bigg| \gamma(x-\tau(x)-u)- \gamma(x-u)\bigg| +  2 \pi C'\big|\langle \tau,x-u \rangle  \gamma(x-u) \big|.\\
    \end{align*}\\
However, $\big|\langle \tau,x-u \rangle  \big| \leq ||\tau||_2||x-u||_2 \leq C''||\tau||_\infty||x-u||_\infty $
Hence, we have\\
$\begin{aligned}
    |k(x,u)| & \leq \bigg| \gamma(x-\tau(x)-u)- \gamma(x-u)\bigg| +  2 \pi D|| \tau||_\infty ||x-u||_\infty |\gamma(x-u)| \\
    &= k_1(x,u)+2 \pi D|| \tau||_\infty k_2(x,u) \quad \text{(say)}.
\end{aligned}$
    \begin{flushleft}
But, from Lemma \eqref{lem:k}, we have 
\begin{equation} \label{eq:k1}
    \sup_{x \in \mathbb{R}^n} \int_{\mathbb{R}^n}| k_1(x,u) | du \leq || \tau||_\infty || \nabla \gamma ||_1.
\end{equation}
On the other hand, by applying change of variables, we can show that\\
\end{flushleft}
\begin{center}
    $$\int_{\mathbb{R}^n}| k_2(x,u) | du = \int_{\mathbb{R}^n} ||y||_\infty |\gamma(y)| dy.$$

\end{center}
Since $\gamma \in S(\mathbb{R}^{n}) $ , we can show that $||y||_\infty^N |\gamma(y)| \leq C_N.$ Further, there exists an $M_N$ such that\\
\begin{equation}
    (1+||y||_\infty^2)^N |\gamma(y)| \leq M_N.
\end{equation}
Furthermore, for $N>n/2$\\
\begin{equation}
\begin{aligned}
    \int_{\mathbb{R}^n} ||y||_\infty |\gamma(y)| dy & =  \int_{\mathbb{R}^n} \frac{(1+||y||_\infty^2)^N ||y||_\infty |\gamma(y)|}{(1+||y||_\infty^2)^N} dy \\
    & \leq M_{N+1} \int_{\mathbb{R}^n} \frac{1}{(1+||y||_\infty^2)^N}dy.
\end{aligned} \label{eq:k2}
\end{equation}
Then, it follows from Lemma \eqref{lem:cor}, \eqref{eq:k1} and \eqref{eq:k2} that $$ \displaystyle \sup_{x \in \mathbb{R}^n} \int_{\mathbb{R}^n} |k(x,u)| du <\infty$$. Similarly, we get $$\displaystyle \sup_{u \in \mathbb{R}^n}\int_{\mathbb{R}^n} |k(x,u)| dx < \infty$$. Then, the result follows from Schur's lemma.
\end{proof}
\vspace{0.25cm}
Under the hypothesis of Proposition \eqref{prop:band}, we have the following
\begin{theorem}
    The feature extractor satisfies\\
    \begin{equation}
        \displaystyle \limsup _{k \rightarrow \infty} \big | \bigl \|\Phi_{\Omega^\mathcal{A}}^{\,k,\mathcal{A}}(F_\tau f)-\Phi_{\Omega^\mathcal{A}}^{\,k,\mathcal{A}}(f)\bigr \| \big| \leq C ||\tau||_\infty||f||_2.
    \end{equation}
\end{theorem}
\begin{proof}
   Appealing to Theorem \eqref{thm:feature d} and Proposition \eqref{prop:band}, we get \\
   
   \begin{equation}
        \big | \bigl \| \Phi_{\Omega^\mathcal{A}}^{\,k,\mathcal{A}}(M_gF_\tau f)-\Phi_{\Omega^\mathcal{A}}^{\,k,\mathcal{A}}(f)\bigr \| \big| \leq  \big | \bigl \| \Phi_{\Omega^\mathcal{A}}^\mathcal{A}(M_gF_{\tau}f)-\Phi^\mathcal{A}_{\Omega^\mathcal{A}}(f)
\bigr \| \big|
\leq
C||\tau||_{\infty}^{\alpha}||f||_{2} \label{eq:1}
   \end{equation}\\
   But $\Phi_{\Omega^\mathcal{A}}^\mathcal{A}$ is a Lipschitz continuous function. Hence
   \begin{equation} \label{eq:2}
        \big | \bigl \| \Phi_{\Omega^\mathcal{A}}^{\,k,\mathcal{A}}(M_gF_\tau f)-\Phi_{\Omega^\mathcal{A}}^{\,k,\mathcal{A}}(F_\tau f)\bigr \| \big| \leq   \big | \bigl \| \Phi_{\Omega^\mathcal{A}}^{\,\mathcal{A}}(M_gF_\tau f)-\Phi_{\Omega^\mathcal{A}}^{\,\mathcal{A}}(F_\tau f)\bigr \| \big| \leq \big |\big | (M_gF_\tau f)-(F_\tau f)\big |\big|.
   \end{equation}

Thus, it follows from \eqref{eq: conv}, that the RHS of \eqref{eq:2} $\rightarrow 0 \quad \text{as} \quad  k \rightarrow \infty$. Hence, from \eqref{eq:1} and \eqref{eq:2}, we have the required result.\\
\end{proof}

\subsection{\textbf{Bounds for cartoon functions}}\mbox{}\\

In this section, we study the deformation insensitivity of cartoon functions. They provide a good model for natural images such as those in the Caltech-256 and
CIFAR-100 data sets, MNIST data set for images of handwritten digits and for images of geometric objects of different shapes, sizes, and colors. 
\begin{definition}
(\cite{article2})
The function \(f : \mathbb{R}^2 \to \mathbb{C}\) is referred to as a \emph{cartoon function} if it can be written as \(f = f_1 + \chi_B f_2\), where \(B \subseteq \mathbb{R}^2\) is a compact domain whose boundary \(\partial B\) is a compact topologically embedded \(C^2\)-hypersurface of \(\mathbb{R}^2\) without boundary, and \(f_i \in L^2(\mathbb{R}^2) \cap C^1(\mathbb{R}^2, \mathbb{C})\), \(i = 1, 2\), satisfy the decay condition
\begin{align*}
|\nabla f_i(x)| \leq C \frac{1}{1+x^2}, \quad i = 1, 2, 
\end{align*}
for some \(C > 0\) (not depending on \(f_1,f_2\)). Furthermore, we denote by
\begin{align*}
\mathcal{C}_{\mathrm{CART}}^K := \{f_1 + \chi_B f_2 \mid {} & f_i \in L^2(\mathbb{R}^1) \cap C^1(\mathbb{R}^1, \mathbb{C}), \ i = 1, 2, \\
& |\nabla f_i(x)| \leq K\frac{1}{1+x^2}, \ \mathrm{vol}^{1}(\partial B) \leq K, \ \|f_2\|_\infty \leq K\}
\end{align*}
the class of cartoon functions of size \(K > 0\).

\end{definition}
Now, we establish the insensitivity for the class of cartoon functions towards deformations. 
\begin{theorem}
Let
$
\Omega^\mathcal{A}=$\{($\Psi_n^\mathcal{A},\mathcal{M}_n^\mathcal{A},P_n^\mathcal{A}):n \in \mathbb{N}$\}$
$
be an admissible $\mathcal{A}$-module-sequence. Then the feature extractor \(\Phi_{\Omega^\mathcal{A}}^\mathcal{A}\) is deformation-stable on the space of cartoon functions \(\mathcal{C}_{\mathrm{CART}}^K\) w.r.t.\ deformations \((F_\tau f)(x) = f(x - \tau(x))\), i.e., for every \(K > 0\), there exists a constant \(C_K > 0\) (that does not depend on \(\Omega^\mathcal{A}\)) such that for all \(f \in \mathcal{C}_{\mathrm{CART}}^K\), and all (possibly non-linear) \(\tau \in C^1(\mathbb{R}^d, \mathbb{R}^d)\) with \(\|\tau\|_\infty < \frac{1}{2}\) and \(\|D\tau\|_\infty \leq \frac{1}{4}\), it holds that
\begin{align*}
\big| \bigl \|\Phi_{\Omega^\mathcal{A}}^\mathcal{A}(F_\tau f) - \Phi_{\Omega^\mathcal{A}}^\mathcal{A}\bigr \| \big|\leq C_K \|\tau\|_\infty^{1/2}. 
\end{align*}
\end{theorem}
\begin{proof}
    A similar proof can be found in \cite{GrohsWB16} and hence is omitted.
\end{proof}
\vspace{0.5cm}
\section{\textbf{Analysis of energy propagation in deep convolutional neural networks}}
\vspace{0.5cm}
In this section, we discuss the problem of the propagation of energy through deep convolutional neural networks. In
particular, we investigate how the energy of the input signal is distributed
and propagated across successive layers of the network through the scattering
propagator. In order to facilitate this analysis, we introduce the notation
$D_{X,p}^{\mathcal{A}}$, associated with an invertible matrix $X$, and study
its interaction with the scattering propagator. The resulting properties
provide a framework for understanding the behavior of the network under the
corresponding $X$-dependent transformations and serve as a foundation for the
subsequent analysis of energy propagation and stability.\\

In this section, we assume that $Tf=|f|$, $\mathcal{M}_k^\mathcal{A}= C_{-\frac{a}{b}}TC_{\frac{a}{b}}$,  $\mathcal{P}_k^\mathcal{A}=Id$ and $s_k=1$ for all $k \in \mathbb{N}.$ Further fix a countable collection $\{g_{\lambda_k};\lambda_k \in \Lambda_k \}$ such that $\{T_s^\mathcal{A} \check{ g_{\lambda_k}} : s \in \mathbb{R}^n, \lambda_k \in \Lambda_k\}$ constitutes a tight semi-discrete frame for \(L^2(\mathbb{R}^n)\) with frame bound $|b_1 \cdot b_2 \cdots b_n| \leq 1$ with the output generating atom being $\varphi_k \in $ $
\{g_{\lambda_{k+1}}:\lambda_{k+1}\in\Lambda_{k+1}\}$ at each $k$-th layer.\\
Under the assumptions mentioned above, we have the following
\begin{proposition}
     For $f \in L^2(\mathbb{R}^n)$, we have\\
    \begin{equation} \label{eq:conservation}
        \displaystyle \sum_{p \in \Lambda^k}||U^\mathcal{A}[p]f||^2=  \displaystyle \sum_{p \in \Lambda^{k+1}}||U^\mathcal{A}[p]f||^2+ \displaystyle \sum_{p \in \Lambda^k}||\Phi_{\Omega^\mathcal{A}}^{k,\mathcal{A}}[p]f||^2, \quad k \in \mathbb{N}.
    \end{equation}
\end{proposition}
\begin{proof}
    We have\\
    \begin{center}
    $\begin{aligned}
        ||f||^2=  \displaystyle \sum_{\lambda_{k+1} \in \Lambda_{k+1} }||f \star_\mathcal{A}g_{\lambda_{k+1} }||^2+ ||f \star_\mathcal{A}\varphi_{k} ||^2.
    \end{aligned}$
    \end{center}
    Hence, we have for $p \in \Lambda^k$
    \begin{center}
        $\begin{aligned}
            ||U^\mathcal{A}[p]f||^2=  \displaystyle \sum_{\lambda_{k+1}  \in \Lambda_{k+1} }||U^\mathcal{A}[p]f \star_\mathcal{A}g_{\lambda_{k+1} }||^2+ ||U^\mathcal{A}[p]f\star_\mathcal{A}\varphi_{k} ||^2.
        \end{aligned}$
    \end{center}
    But $||\mathcal{M}_k^\mathcal{A}f||=||f||$ for all  $k \in \mathbb{N}.$ Thus,
    \begin{center}
        $\begin{aligned}
            ||U^\mathcal{A}[p]f||^2=  \displaystyle \sum_{\lambda_{k+1}  \in \Lambda_{k+1} }||\mathcal{M}_{k+1}^\mathcal{A}(U^\mathcal{A}[p]f \star_\mathcal{A}g_{\lambda_{k+1} })||^2+ ||U^\mathcal{A}[p]f\star_\mathcal{A}\varphi_{k} ||^2.
            \end{aligned}$\\
            \end{center}
Hence, summing over all paths of length $k,$ we get \\
\begin{center}

        $\begin{aligned}
            \displaystyle \sum_{p \in \Lambda^k}||U^\mathcal{A}[p]f||^2=\displaystyle \sum_{p \in \Lambda^k}\sum_{\lambda_{k+1}  \in \Lambda_{k+1} }||\mathcal{M}_{k+1}^\mathcal{A}(U^\mathcal{A}[p]f \star_\mathcal{A}g_{\lambda_k})||^2+\displaystyle \sum_{p \in \Lambda^k}||\Phi_{\Omega^\mathcal{A}}^{k,\mathcal{A}}[p]f||^2,  
        \end{aligned}$
\end{center}
    from which \eqref{eq:conservation} follows.
    
\end{proof}
\vspace{0.5cm}
\begin{remark}
    The above proposition establishes that, at each layer of the neural network, the energy of the input signal is decomposed into two components: the energy extracted by the scattering coefficients at the current layer and the energy propagated to the subsequent layer. Thus, the total energy is preserved across successive layers of the network. This layer-wise energy conservation provides an important structural property of the scattering architecture. A more general formulation of this energy conservation principle is presented in the following result.
\end{remark}
Let $W_N^\mathcal{A}(f)=\displaystyle \sum_{p \in \Lambda^N} ||U^\mathcal{A}[p]f||^2.$ (Occasionally, we write \(W_N[\Psi^{\mathcal{A}}]f\) to indicate explicitly the dependence of the semi-discrete frame \(\Psi^{\mathcal{A}}\).) Under the assumptions mentioned above, we have the following
\begin{proposition}
    For $f \in L^2(\mathbb{R}^n)$
    \begin{equation}
        ||f||^2=\displaystyle \sum_{n=0}^N \big| \bigl\|\Phi_{\Omega^\mathcal{A}}^{n,\mathcal{A}}f\bigr \| \big|^2+ W_N^\mathcal{A}(f).
    \end{equation}
    
\end{proposition}
\begin{proof}
    The proof is similar to that of Proposition 1 of Appendix of \cite{MR3819343}, so, it is omitted. 
\end{proof}
\begin{lemma}
    For $f,g \in L^2(\mathbb{R}^n)$ and $N \in \mathbb{N}$,
    \begin{enumerate}
        \item [(i)] $|W_N^\mathcal{A}(f)-W_N^\mathcal{A}(g)| \leq \sqrt{2}\cdot||f-g|| \cdot \sqrt{W_N^\mathcal{A}(f)+W_N^\mathcal{A}(g)},\\$
        \item [(ii)] $W_N^\mathcal{A}(f) \geq \frac{1}{2}\cdot W_N^\mathcal{A}(g)-||f-g||^2.$
    \end{enumerate}
\end{lemma}
\begin{proof}
\begin{enumerate}
    \item[(i)] Consider
    \begin{align*}
    |W_N^\mathcal{A}(f)-W_N^\mathcal{A}(g)|^2
      &= \bigg| \sum_{p \in \Lambda^n}\|U^\mathcal{A}[p]f\|^2-\sum_{p \in \Lambda^n}\|U^\mathcal{A}[p]g\|^2 \bigg|^2\\
      &= \bigg| \sum_{p \in \Lambda^n}\big(\|U^\mathcal{A}[p]f\|-\|U^\mathcal{A}[p]g\|\big)\,\big(\|U^\mathcal{A}[p]f\|+\|U^\mathcal{A}[p]g\|\big) \bigg|^2\\
      &\le \left(\sum_{p \in \Lambda^n}\big(\|U^\mathcal{A}[p]f\|-\|U^\mathcal{A}[p]g\|\big)^2\right)
            \left(\sum_{p \in \Lambda^n}\big(\|U^\mathcal{A}[p]f\|+\|U^\mathcal{A}[p]g\|\big)^2\right)\\
      &\le \left(\sum_{p \in \Lambda^n}\|U^\mathcal{A}[p]f-U^\mathcal{A}[p]g\|^2\right)
            \left(\sum_{p \in \Lambda^n}\big(\|U^\mathcal{A}[p]f\|+\|U^\mathcal{A}[p]g\|\big)^2\right),
            \end{align*}\\
            by using Schwarz inequality. Using the fact that $(a+b)^2 \leq 2(a^2+b^2)$ and \eqref{eq:lipschitz}, we get\\
      \begin{center}      
     $|W_N^\mathcal{A}(f)-W_N^\mathcal{A}(g)|^2 \le 2\,\|f-g\|^2\,\big(W_N^\mathcal{A}(f)+W_N^\mathcal{A}(g)\big).$
   \end{center}

    \item[(ii)] On the other hand,\\
    \begin{align*}
        W_N^\mathcal{A}(g)&=\displaystyle \sum_{p \in \Lambda^n}||U^\mathcal{A}[p]g||^2\\
        & \leq \displaystyle \sum_{p \in \Lambda^n}\bigg(||U^\mathcal{A}[p](g-f)||+ ||U^\mathcal{A}[p]f||\bigg)^2\\
        & \leq 2 \displaystyle \sum_{p \in \Lambda^n}||U^\mathcal{A}[p](g-f)||^2+ 2\displaystyle \sum_{p \in \Lambda^n}||U^\mathcal{A}[p]f||^2\\
        & \leq 2 ||f-g||^2 + 2 W_N^\mathcal{A}(f),
    \end{align*}
    from which the result follows.

\end{enumerate}
\end{proof}

\begin{definition}
    Let $X$ be an invertible matrix of order $n.$ Let $f:\mathbb{R}^n \rightarrow \mathbb{C}$. Define $D_X^p(f):=|\det(X)|^{1/p}f(X\,\cdot)$ for $1 \leq p < \infty$. We define the dilation matrix associated with the SAFT and the matrix $X$ as\\
    \begin{equation}
        D_{X,p}^\mathcal{A}:= C_{- \frac{a}{b}}D_X^pC_{ \frac{a}{b}}.
    \end{equation}
\end{definition}
\begin{remark}
    Let $Tf=|f|.$ Since $T$ commutes with $D_X^p$, $M=C_{- \frac{a}{b}}TC_{ \frac{a}{b}}$ commutes with $ D_{X,p}^\mathcal{A}$. 
\end{remark}
\vspace{0.25cm}
\begin{remark}\label{rem:fuhr}
    For $p=1,$ it is a straight forward calculation to see that\\
    \begin{center}
        $ D_X^1f \star g
        =
        D_X^1
        \left(
        f \star
        \left(D_{X^{-1}}^1g\right)
        \right),$
    \end{center}
    for $f,g \in L^2(\mathbb{R}^n)$.
\end{remark}
\begin{proposition}
 Let $\{g_{\lambda_k}:\lambda_k \in \Lambda_k \}$ be a countable collection  such that $\Psi_k^\mathcal{A}=\{T_s^\mathcal{A} \check{g_{\lambda_k}} : s \in \mathbb{R}^n, \lambda_k \in \Lambda_k\}$ constitutes a semi-discrete frame for \(L^2(\mathbb{R}^n)\).  Further, assume that $D_{X^{-1},1}^\mathcal{A}\Psi^\mathcal{A}:=\{T_s^\mathcal{A}\check{ D_{X^{-1},1}^\mathcal{A}g_{\lambda_k}} : s \in \mathbb{R}^n, \lambda_k \in \Lambda_k\} $ forms a semi-discrete frame for \(L^2(\mathbb{R}^n)\). For $N \in \mathbb{N}, \quad q \in \Lambda^N,\quad X \in GL_n(\mathbb{R)}$ and $f \in L^2(\mathbb{R}^n)$, let $D_{X^{-1},1}^\mathcal{A}\lambda_j=D_{X^{-1},1}^\mathcal{A}g_{\lambda_j}$.\\
    \begin{enumerate}
        \item[(i)] Let $ D_{X^{-1},1}^\mathcal{A}[q] :=
\left(D_{X^{-1},1}^\mathcal{A}\lambda_1,\ldots,D_{X^{-1},1}^\mathcal{A}\lambda_N\right).$ Then \\
\begin{center}
$U^\mathcal{A}[q]D_{X,1}^\mathcal{A}f=D_{X,1}^\mathcal{A}U^\mathcal{A}[D_{X^{-1},1}^\mathcal{A}q]f$.
\end{center}
\item[(ii)] Moreover, if $f \in L^2(\mathbb{R}^n)$,\\
\begin{center}
$W_N[\Psi^\mathcal{A}](D_{X,2}^\mathcal{A}f)=W_N[D_{X^{-1},1}^\mathcal{A}\Psi^\mathcal{A}](f).$
\end{center}
    \end{enumerate}
\end{proposition}
\begin{proof}

\begin{enumerate}
    \item[(i)] We have, for all $n \in \mathbb{N},$
    \begin{align}
        D_X^1f \star g_{\lambda_n}
        &=
       D_X^1
        \left(
        f \star
        \left(D_{X^{-1}}^{1}g_{\lambda_n}\right)
        \right).
    \end{align}
    using Remark \eqref{rem:fuhr}. Consider \\
    \begin{align}
        D_{X,1}^\mathcal{A}f \star_\mathcal{A}g_{\lambda_n}&=\sqrt{|b_1 \cdot b_2 \cdots b_n|}C_{-\frac{a}{b}}(C_{\frac{a}{b}}D_{X,1}^\mathcal{A}f \star C_{\frac{a}{b}} g_{\lambda_n})&&\notag\\
        &=\sqrt{|b_1 \cdot b_2 \cdots b_n|}C_{-\frac{a}{b}} D_X^1(C_{\frac{a}{b}}f \star D_{X^{-1}}^1 C_{\frac{a}{b}}g_{\lambda_n})&&\notag\\
        &= C_{- \frac{a}{b}}D_X^1C_{ \frac{a}{b}}(f \star_\mathcal{A}  C_{- \frac{a}{b}}D_{X^{-1}}^1C_{ \frac{a}{b}})&&\notag\\
        &= D_{X,1}^\mathcal{A}(f \star_\mathcal{A} D_{X^{-1},1}^\mathcal{A}g_{\lambda_n}).
    \end{align}\\
    But \\
    \begin{align}
        U^{\mathcal{A}}[g_{\lambda_n}]
        D_{X,1}^{\mathcal{A}}f
        &=
        D_{X,1}^{\mathcal{A}}
        U^{\mathcal{A}}
        \left[D_{X^{-1},1}^{\mathcal{A}}g_{\lambda_n}\right]f.
    \end{align}
    Hence, by iterating this relation along the path $q$, we obtain the
    desired result.

    \item[(ii)] Since $D_X^1=|\det(X)|^{1/2} D_X^2,$ we get\\
    \begin{align}
        D_{X,1}^{\mathcal{A}}
        &=
        |\det(X)|^{1/2}D_{X,2}^{\mathcal{A}}.
    \end{align}
    Then, it follows from (i) that
    \begin{align}
        U^{\mathcal{A}}[q]D_{X,2}^{\mathcal{A}}
        &= |\det(X)|^{-1/2}D_{X,1}^{\mathcal{A}}U^{\mathcal{A}}
        \left[D_{X^{-1},1}^{\mathcal{A}}q\right]\\
         &=D_{X,2}^{\mathcal{A}}
        U^{\mathcal{A}}
        \left[D_{X^{-1},1}^{\mathcal{A}}q\right].
    \end{align}
    Since $D_{X,2}^{\mathcal{A}}$ is a unitary operator on
    $L^2(\mathbb{R}^n)$, summing over all paths in $\Lambda^N$ gives the
    desired result.
\end{enumerate}
\end{proof}
\vspace{0.5cm}   
\section{\textbf{DCNN based on DSAFT and its application to digit classification}}
\vspace{0.5cm}
We now introduce the deep convolutional network associated with DSAFT.\\
 \begin{definition}
     Consider $\mathbb{Z}_m=\{ [0],[1], \cdots [m-1] \} $, the integers modulo $m$ and take $L_m= \sqrt{|b|}\mathbb{Z}_m$ where $b$ is a non-zero real number. Then the square lattice $L_m \times L_m$ forms an abelian group with the binary operation given below:\\
     
     $\text{For} \quad l=(l_1,l_2),l'=(l_1',l_2') \in L_m \times L_m, $\\
     \begin{center}
         $\begin{aligned}
          l+l':=\bigg(\sqrt{|b|}\Big(\frac{l_1+l_1'}{\sqrt{|b|}}\mod  m \Big),\sqrt{|b|}\bigg(\frac{l_2+l_2'}{\sqrt{|b|}}\mod  m \Big) \bigg)
     \end{aligned}$
     \end{center}
     We consider the space of m periodic functions $l^2(L_m \times L_m) $ with the associated  inner product\\
     \begin{center}
         $\begin{aligned}
        \langle f, f' \rangle
=
\frac{b^2}{m^2}\sum_{n_1 \in L_m}
\sum_{n_2 \in L_m}
f(n_1,n_2)\overline{f'(n_1,n_2)},
\qquad
f,f' \in \ell^2(L_m \times L_m).
\end{aligned}$
     \end{center}
 \end{definition}
 \vspace{0.5cm}
 Now we define the discrete SAFT on $l^2(L_m \times L_m) $.\\
 \begin{definition}
     Let $f \in l^2(L_m \times L_m)$ and $a,b,c,d,p,q \in \mathbb{R}$ with $b>0$, the \emph{discrete special affine Fourier transform (DSAFT)} is defined as follows:\\
     \begin{equation}
\begin{aligned}
\mathcal{F}_{\mathcal{A}}f(n)
&= \frac{1}{b}\displaystyle\sum_{k\in L_m\times L_m} f(k)
\exp\bigg[
\frac{\pi i}{b}
\bigg(
a\big([k_1^2+k_2^2]\big)
-2\bigg(\frac{[k_1n_1]}{m}-[k_1]p\bigg) \\
&\qquad\qquad
-2\bigg(\frac{[k_2n_2]}{m}-[k_2]p\bigg)
\bigg)
\bigg] \\
&\qquad\qquad\times
\exp\bigg[
-2\bigg(\frac{[n_1+n_2]}{m}\bigg)(dp-bq)
+d\bigg(\frac{[n_1^2+n_2^2]}{m^2}\bigg)
\bigg], , \quad n \in L_m \times L_m.
\end{aligned}
\end{equation}
 For notational convenience, we write \\
 \begin{equation}
\begin{aligned}
K_{\mathcal{A}}(k,n)
&= \frac{1}{b}\exp\bigg[
\frac{\pi i}{b}
\bigg(
a\big([k_1^2+k_2^2]\big)
-2\bigg(\frac{[k_1n_1]}{m}-[k_1]p\bigg) \\
&\qquad\qquad
-2\bigg(\frac{[k_2n_2]}{m}-[k_2]p\bigg)
\bigg)
\bigg] \\
&\qquad\times
\exp\bigg[
-2\bigg(\frac{[n_1+n_2]}{m}\bigg)(dp-bq)
+d\bigg(\frac{[n_1^2+n_2^2]}{m^2}\bigg)
\bigg].
\end{aligned}
\end{equation}\\
Here, $[k_i]$ and $[n_i]$ denote the equivalence classes of $k_i$ and $n_i$, respectively, in $L_m$. Hence, the special affine Fourier transform also belongs to $l^2(L_m \times L_m).$ For simplicity, we omit the equivalence class notation. 

 \end{definition}

 \begin{lemma}
  DSAFT is an isometry on $l^2(L_m \times L_m)$.   
 \end{lemma} 
 \begin{proof}
 Consider\\
  \begin{align*}
         \langle \mathcal{F}_\mathcal{A}f,\mathcal{F}_\mathcal{A}f \rangle 
         &=\frac{b^2}{m^2} \displaystyle \sum_{r \in L_m \times L_m} \sum_{s \in L_m \times L_m} f(s) K_\mathcal{A}(s,r)\sum_{t \in L_m \times L_m} \overline{ f(t) K_\mathcal{A}(t,r)}.\\
         \end{align*}
Taking $r=\sqrt{b}x,s=\sqrt{b}y$ and $t=\sqrt{b}z$,
\begin{align*}
          \langle \mathcal{F}_\mathcal{A}f,\mathcal{F}_\mathcal{A}f \rangle &=\frac{b^2}{m^2} \displaystyle \sum_{x \in \mathbb{Z}_m \times\mathbb{Z}_m} \sum_{y \in\mathbb{Z}_m\times \mathbb{Z}_m} f(\sqrt{b}y) K_\mathcal{A}(\sqrt{b}y,\sqrt{b}x)\sum_{z \in \mathbb{Z}_m \times \mathbb{Z}_m} \overline{ f(\sqrt{b}z) K_\mathcal{A}(\sqrt{b}z,\sqrt{b}x)}.\\
          \end{align*}
On simplification, we get\\
\begin{align*}
          \langle \mathcal{F}_\mathcal{A}f,\mathcal{F}_\mathcal{A}f \rangle
          &=\frac{1}{m^2}\sum_{y \in\mathbb{Z}_m\times \mathbb{Z}_m} \sum_{z \in\mathbb{Z}_m\times \mathbb{Z}_m} f(\sqrt{b}y)\overline{f(\sqrt{b}z)}\exp\bigg( \pi i (||y||^2-||z||^2) \bigg)\\
         & \quad \exp \bigg(\frac{2 \pi i}{\sqrt{b}}(y_1+y_2-z_1-z_2)p \bigg)
          \sum_{x \in\mathbb{Z}_m\times \mathbb{Z}_m} \exp \bigg( \frac{-2 \pi i(x\cdot (y-z))}{m}\bigg).\\
        \end{align*}\\
But,\\
$\displaystyle \sum_{x \in\mathbb{Z}_m\times \mathbb{Z}_m} \exp \bigg( \frac{-2 \pi i(x\cdot (y-z))}{m}\bigg)= \begin{cases}
    m^2 &,y=z\\ 
    0&, y \neq z.
\end{cases}$\\
Thus,\\
\begin{align*}
          \langle \mathcal{F}_\mathcal{A}f,\mathcal{F}_\mathcal{A}f \rangle
          &= \frac{1}{m^2}\sum_{y \in\mathbb{Z}_m\times \mathbb{Z}_m}  f(\sqrt{b}y)\overline{f(\sqrt{b}y)}m^2\\
          &=\langle f,f \rangle .
        \end{align*}\\
     proving our assertion.
 \end{proof}    
     \vspace{0.5cm}
     Now, we define the translation, modulation and convolution w.r.t discrete SAFT. 
     \begin{definition}
         For $s \in L_m \times L_m, $ $  f \in l^2(L_m \times L_m),$ we define the $\mathcal{A}$-\emph{translation} of $f$ by $s$ as follows:\\
         \begin{equation}
             T_s^\mathcal{A}f(t):=\exp\bigg(-2 \pi i \frac{a}{b}\displaystyle \sum_{j=1}^2 s_j(t_j-s_j) \bigg) f(t-s), \quad t \in L_m \times L_m. 
         \end{equation}
     \end{definition}
     \begin{definition}
       For $s \in L_m \times L_m,$ $ f \in l^2(L_m \times L_m),$ we define the $\mathcal{A}$-\emph{modulation} of $f$ by $s$ as follows:\\
       \begin{equation}
         M_s^\mathcal{A}(f)(t):=\exp \bigg(\pi i \displaystyle \sum_{j=1}^2 \frac{a\frac{s_j^2}{m^2}-\frac{2ps_j}{m}+\frac{2s_jt_j}{m}}{b} \bigg)f(t), \quad t \in L_m \times L_m.
         \end{equation}
     \end{definition}
     \begin{remark}
         We have, \\
         \begin{equation}
             \mathcal{F_A}(T_s^\mathcal{A}f)(n)=\exp \bigg( \pi i\frac{a( s_1^2+s_2^2)}{b}(1-\frac{1}{m^2}) + 2 \pi i p\big(\frac{s_1+s_2}{b}\big)(1-\frac{1}{m})\bigg)M_{-s}^\mathcal{A}(\mathcal{F_A}f).
         \end{equation}
     \end{remark}
     \begin{definition}
         For $f,g \in l^2(L_m \times L_m),$ we define the $\mathcal{A}$-\emph{convolution} of $f$ and $g$ by\\
         \begin{equation}
             f \star_\mathcal{A}g(n):= \displaystyle \sum_{m \in L_m \times L_m}f(m) T_m^\mathcal{A}g(n), \quad n \in L_m \times L_m.
         \end{equation}
     \end{definition}
     \begin{definition}
         For $a,b \in \mathbb{R}$, $f \in l^2(L_m \times L_m)$ we define the \emph{chirp modulation} operator as\\
         \begin{equation}
        C_{\frac{a}{b}}f(t):=\exp\bigg(\pi i\displaystyle \frac{a}{b} \sum_{j=1}^2 t_j^2 \bigg)f(t), \quad t \in L_m \times L_m.
    \end{equation}
    \end{definition}
    \begin{remark}
    As in the continuous setting, \eqref{eq:chirp conv} and \eqref{eq:chirp trans} continue to hold in the discrete case. In fact, we have $|\mathcal{F}_\mathcal{A}(f\star_\mathcal{A}g)|=|\mathcal{F}_\mathcal{A}f||\mathcal{F}_\mathcal{A}g|.$
   \end{remark}

      \begin{definition}
For $\Lambda$, a countable index set, the family $\Psi_\Lambda^\mathcal{A}=\{g_\lambda: \lambda \in \Lambda \}$ is called an $\mathcal{A}$- convolutional set with Bessel bound $B>0$ if\\
\[
\displaystyle \sum_{\lambda \in \Lambda} ||f \star_\mathcal{A} g_\lambda||^2 \leq B||f||^2, \quad f \in l^2(L_m \times L_m).
\]
\end{definition}  
\begin{remark}
One can obtain similar results for the $\mathcal{A}$-convolutional set as in the case of Proposition \eqref{prop:equivalence}.
\end{remark}

\vspace{0.5cm}
    
Now, we discuss the DCNN architecture associated with the DSAFT. \\
\vspace{0.25cm}

For $N \in \mathbb{N}$, let $H_N=l^2(L_N\times L_N)$. Consider a network with \(D\) layers. For each layer \(d\), \(1 \leq d \leq D\), let $
\Psi_d^{\mathcal{A}} = \{ g_{\lambda_d}:\lambda_d \in \Lambda_d\} \subseteq H_{N_d}$
be an $\mathcal{A}$-convolutional set,  
$\rho_d^{\mathcal{A}} : \mathbb{C} \times\mathbb{C} \to \mathbb{C}\times \mathbb{C}$, a pointwise Lipschitz-continuous nonlinearity and  
$P_d^{\mathcal{A}} : H_{N_d} \to H_{N_{d+1}}
$, a Lipschitz-continuous pooling operator with \(N_{d+1} = \frac{N_d}{S_d}\), where \(S_d \in \mathbb{N}\) denotes the pooling factor in the \(d\)-th layer. The parameters of these three components are allowed to vary over all layers. Thus, the \(d\)-th layer is characterized by the triplet 
$\Omega_d^{\mathcal{A}} := \left( \Psi_d^{\mathcal{A}}, \rho_d^{\mathcal{A}}, P_d^{\mathcal{A}} \right).$
Notice that the dimensions of the spaces \(H_{N_d}\) satisfy 
$N_1 \geq N_2 \geq \cdots \geq N_D.$
The collection 
$\Omega^{\mathcal{A}} := \left\{ \Omega_d^{\mathcal{A}} \right\}_{d=1}^{D}
$ will be referred to as the $\mathcal{A}$-module-sequence of the network.

For each $\lambda_d\in\Lambda_d$, the action of the $d$-th $\mathcal{A}$-module on an input feature map $f$ is defined as follows:
\begin{center}
$U_d^\mathcal{A}[\lambda_d]f=P_d^\mathcal{A}\left(\rho_d^\mathcal{A}(f*_\mathcal{A}g_{\lambda_d})\right).$
\end{center}
Hence, each layer produces a family of feature maps indexed by the elements of $\Lambda_d$. These feature maps are subsequently propagated to the next layer and serve as inputs for further filtering and nonlinear processing.

The successive application of layer operators gives rise to different paths through the network. For a path
$q=(\lambda_1,\lambda_2,\ldots,\lambda_d)
\in\Lambda_1\times\Lambda_2\times\cdots\times\Lambda_d,$
the corresponding propagated feature is obtained by
\begin{center}
$
U^\mathcal{A}[q]f=U_d^\mathcal{A}[\lambda_d]\cdots U_2^\mathcal{A}[\lambda_2]U_1^\mathcal{A}[\lambda_1]f.
$
\end{center}
The empty path $e$ is identified with the input itself, i.e.,$U^\mathcal{A}[e]f=f.$ Thus, a path records the sequence of filters selected at successive layers and describes the hierarchical transformation of the input signal through the network.

In order to construct the final feature representation, the output may be extracted from one or more layers. For a path $q$ of length $d$, the corresponding output is obtained by applying an output-generating filter $\chi_d \in \{ g_{\lambda_{d+1}}:\lambda_{d+1} \in \Lambda_{d+1}\}$ to the propagated feature,
$
(U^\mathcal{A}[q]f)*_\mathcal{A}\chi_d.$
The choice of $\chi_d$ determines the contribution of the $d$-th layer to the feature representation. In particular, taking $\chi_d$ to be an identity-type filter allows the propagated feature itself to be retained, whereas setting $\chi_d=0$ excludes that layer from the final representation.

Accordingly, the feature extractor associated with the $\mathcal{A}$-module sequence $\Omega^\mathcal{A}$ is formed by collecting the selected outputs over the admissible paths and network layers. This construction allows information from different depths to be incorporated into the feature representation while retaining the hierarchical structure of the network.\\

Now, we establish the global properties of the feature
vectors generated by the proposed network. In particular, we study the
Lipschitz continuity and stability of the feature extractor and derive
bounds describing its behavior across the different layers of the
network.

\begin{theorem}
Let $
\Omega^\mathcal{A}
=
\bigl\{
(\Psi_d^\mathcal{A},P_d^\mathcal{A},\rho_d^\mathcal{A})
: 1\leq d\leq D\bigr \}$ be an $\mathcal{A}$-module-sequence. Assume that the Bessel bounds
$B_d^\mathcal{A}>0$, the Lipschitz constants $L_d^\mathcal{A}>0$ of the
nonlinearities $\rho_d^\mathcal{A}$, and the Lipschitz constants
$R_d^\mathcal{A}>0$ of the pooling operators $P_d^\mathcal{A}$ satisfy
\begin{equation} \label{eq:adms}
\max_{1\leq d\leq D}
\max\left\{
B_d^\mathcal{A},
B_d^\mathcal{A}(R_d^\mathcal{A})^2(L_d^\mathcal{A})^2
\right\}
\leq 1.
\end{equation}

Then the following properties hold.

\begin{enumerate}
    \item[(i)] The feature extractor $\Phi_{\Omega^\mathcal{A}}^\mathcal{A}$ is
    Lipschitz-continuous with Lipschitz constant
    $L_{\Omega}^\mathcal{A}=1$, i.e.,
    \begin{align*}
   \big| \bigl \|
    \Phi_{\Omega^\mathcal{A}}^\mathcal{A}(f)
    -\Phi_{\Omega^\mathcal{A}}^\mathcal{A}(h)
   \bigr \| \big|
    \leq
    \|f-h\|_2,
    \end{align*}
    for all $f,h\in H_{N_1}$, where the norm on the feature space is
    defined by
    \begin{align*}
    \big| \bigl \|
    \Phi_{\Omega^\mathcal{A}}^\mathcal{A}(f)
    \bigr \| \big|^2
    :=
    \sum_{d=0}^{D-1}
    \sum_{q\in\mathcal{A}^d}
    \left\|
    U^\mathcal{A}[q]f*_\mathcal{A}\chi_d
    \right\|_2^2.
    \end{align*}

    \item[(ii)] In addition to condition \eqref{eq:adms}, suppose that, for
    every $d\in\{1,\ldots,D-1\}$, the nonlinearities
    $\rho_d^\mathcal{A}$ and the pooling operators $P_d^\mathcal{A}$
    satisfy
    \[
    \rho_d^\mathcal{A}(0)=0,
    \qquad
    P_d^\mathcal{A}(0)=0.
    \]
    Then
    \begin{align*}
    \big| \bigl \|
    \Phi_{\Omega^\mathcal{A}}^\mathcal{A}(f)
    \bigr \| \big|
    \leq
    \|f\|_2,
    \qquad
    \forall f\in H_{N_1}.
    \end{align*}

\end{enumerate}
\end{theorem}
\begin{proof}
    A similar proof is given in Appendix E of \cite{pmlr-v48-wiatowski16} and hence is omitted.
\end{proof}

\begin{table}[!t]
\centering
\caption{Classification results on the MNIST dataset for the proposed scattering-based representation and the baseline FCN.}
\label{tab:MNIST}
{\small
\begin{tabular}{|c|c|c|c|}
\hline
$a$ & $b$ & Corresponding transform & Accuracy \\
\hline
$\sqrt{2}$ & $1$ & Linear canonical transform & $99\%$ \\[1mm]
$\cos(\pi/3)$ & $\sin(\pi/3)$ & Fractional Fourier transform & $98.7\%$ \\[1mm]
$1$ & $2/3$ & Fresnel transform & $98.8\%$ \\[1mm]
$0$ & $1$ & Fourier transform & $99.2\%$ \\[1mm]
\hline
\multicolumn{3}{|c|}{Raw pixel representation (baseline)} & $98\%$ \\[1mm]
\hline
\end{tabular}
}
\end{table}
\subsection{\textbf{Numerical implementation using Python}}\mbox{}\\

We consider the problem of handwritten digit classification to evaluate the effectiveness of the feature extractor $\Phi_{\Omega^\mathcal{A}}^\mathcal{A}$ when coupled with a fully connected neural network (FCN) classifier. The experiments are conducted on the MNIST dataset, which consists of $60,000$ training images and $10,000$ test images, each of size $28\times28$. The objective is to investigate the discriminative capability of the feature representations generated by the proposed scattering-based architecture for handwritten digit recognition.

We construct a discrete-time deep convolutional network with $D=3$ layers. In each layer, the same type of filters, pointwise nonlinearity is employed. The $\mathcal{A}$-convolutional set underlying the network is constructed in a manner analogous to the construction presented in Example \eqref{ex:2D}. In the present setting, the Haar wavelet is used to construct it, which subsequently determines the collection of filters employed in the convolutional operations. The wavelet filters are generated over $J=3$ scales and are sensitive to the three principal directions, namely, horizontal, vertical, and diagonal directions. At each layer, the convolutional responses are transformed using the absolute-value nonlinearity, followed by low-pass filtering to obtain the corresponding scattering features. Circular convolution is used in the implementation of the convolutional operations.

The absolute-value operator is employed as the pointwise nonlinearity throughout the network. Following the convolutional operations and nonlinear transformation, max pooling is applied to progressively reduce the spatial resolution of the feature maps. The feature representations obtained from the successive layers are retained and subsequently combined to form the final scattering representation. This hierarchical construction enables the extracted features to incorporate information across multiple scales while progressively reducing the spatial resolution.

The resulting scattering features are subsequently provided as input to a fully connected neural network (FCN) for classification. The feature maps generated by the scattering network are flattened and supplied to the FCN, whose parameters are learned using the training data. The FCN performs the final classification into the ten digit classes of the MNIST dataset. For comparison, a conventional FCN is also trained directly on the input images, providing a baseline against which the effectiveness of the proposed scattering-based feature representation can be assessed.

The classification performance of both approaches is evaluated on the MNIST test set. The corresponding results are presented in Table \eqref{tab:MNIST}. The comparison allows us to assess the contribution of the proposed scattering-based feature extraction stage to the subsequent FCN classification and to examine whether the multiscale representation obtained through the proposed architecture provides improved discriminative information compared with direct classification using an FCN.

\nocite{*}

\bibliographystyle{amsplain} 
\bibliography{references}
\end{document}